\pdfoutput=1
\documentclass{article}

\usepackage{xcolor}
\usepackage[preprint]{corl_2026}

\usepackage{url}

\usepackage{booktabs}
\usepackage{tabularx}
\usepackage{graphicx}
\usepackage{xspace}
\usepackage{makecell}
\usepackage{algorithm}
\usepackage{algpseudocode}
\usepackage{multirow}
\usepackage{subfig}
\usepackage{tcolorbox}
\tcbuselibrary{breakable}
\usepackage{bbm}
\usepackage{tikz}
\usepackage{amssymb}
\usepackage{wrapfig}

\newcounter{prompt}
\renewcommand{\theprompt}{\Alph{prompt}}

\usepackage{mathtools}

\newcommand{\figref}[1]{Fig.~\ref{#1}}
\newcommand{\secref}[1]{Section~\ref{#1}}
\newcommand{\tabref}[1]{Table~\ref{#1}}
\newcommand{\apref}[1]{Appendix~\ref{#1}}

\newcommand{\equationref}[1]{Eq.~\ref{#1}}

\newcommand{\expr}{e}
\newcommand{\obj}{r}
\newcommand{\step}{t}
\newcommand{\mstep}{T}

\newcommand{\gt}{y}
\newcommand{\pred}{\hat{\gt}}

\newcommand{\gts}{\mathcal{Y}}
\newcommand{\preds}{\hat{\gts}}
\newcommand{\tspc}{\mathcal{T}}
\newcommand{\ispc}{\mathcal{I}}
\newcommand{\rspc}{\mathcal{R}}

\newcommand{\act}{\mathcal{A}}
\newcommand{\acti}{\act_q}
\newcommand{\actc}{\act_c}
\newcommand{\actg}{\act_g}
\newcommand{\actr}{\act_r}

\newcommand{\testit}[1]{\noindent\textbf{\textit{#1}}\ }

\definecolor{mint}{rgb}{0.24, 0.71, 0.54}

\newcommand{\hanbo}[1]{#1}
\newcommand{\tred}[1]{#1}
\newcommand{\bst}[1]{}

\newlength{\paragap}
\newcommand{\pgap}{\vspace{\paragap}}

\title{Surrogate Uncertainty for Black-box Visual Grounding}

\author{
  Hanbo Zhang$^{1,2,}$\thanks{Work mainly done as a postdoctoral researcher at the AdaComp Lab, National University of Singapore.}\hphantom{$^{*}$}, Tianrun Hu$^{1}$, Cedar Site Bai$^{3}$, Anxing Xiao$^{1}$, David Hsu$^{1}$ \\
  $^1$National University of Singapore\\
  $^2$Shanghai Innovation Institute\\
  $^3$Purdue University
}

\begin{document}
\maketitle

\vspace{-0.2cm}
\begin{abstract}
Visual grounding is crucial for robots to understand language instructions with visual observations.
This paper addresses the problem of visual grounding uncertainties, where a robot uses a black-box vision-language model (VLM) to ground free-form natural-language instructions to target objects. Prevailing methods derive uncertainty from the VLM itself. Yet, outputs of the VLM, e.g., generated uncertainties, are from the same parameters and structures producing the predictions, and inherit the bias. We give a formal Bayesian definition of the problem that treats the VLM prediction as an observation. An external surrogate Bayesian model then quantifies per-query uncertainty over a local neighborhood of a held-out calibration set, decomposed into aleatoric and epistemic components. A hierarchical retrieval across input modalities realizes a compact joint-space neighborhood, mitigating the curse of dimensionality. The estimator further integrates with conformal prediction or model-based planning for active visual grounding, where robots are allowed to take verbal or physical actions to ground targets. Across grounding benchmarks and state-of-the-art VLMs, it consistently improves both calibration and task performance. On a real Fetch robot under open-world instructions, it enables principled trade-offs between confident action and information gathering, substantially improving robustness.
\end{abstract}


\section{Introduction}

\begin{wrapfigure}{r}{0.48\textwidth}
\vspace{-1.2cm}
\centering
\includegraphics[width=0.46\textwidth]{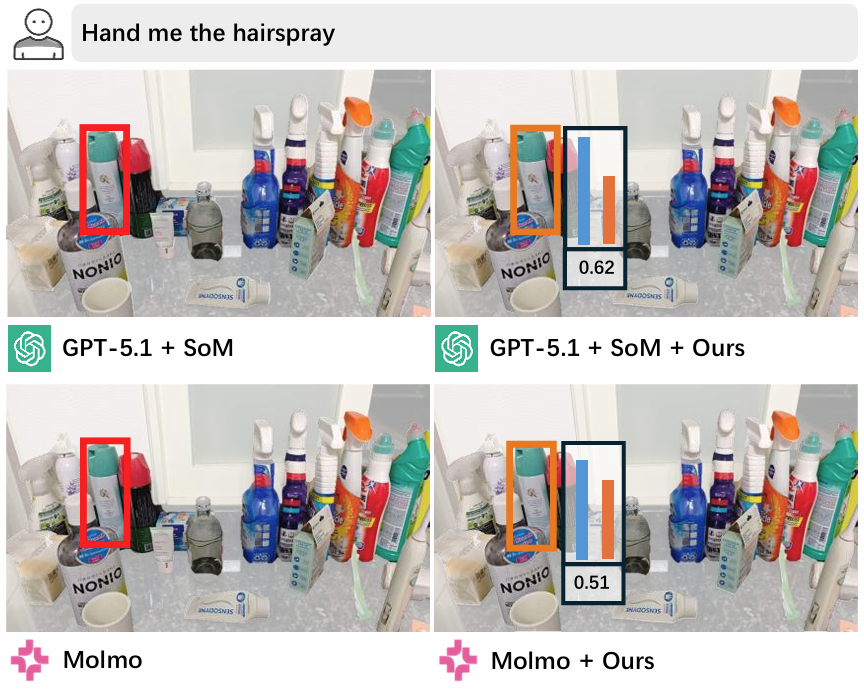}
\caption{An example of uncertainty estimation, where uncertainty matters. Without uncertainty, a dangerous wrong object will be selected by the robot (e.g., take ``disinfectant'' as ``hairspray'').\tred{\ Red boxes show predictions. Bars show the grounding probabilities of negative (orange) and positive (blue) from our estimated uncertainty.}}
\label{fig:intro}
\vspace{-0.5cm}
\end{wrapfigure}
Reliable uncertainty estimation is crucial for robots acting in physical environments. In safety-critical scenarios, overconfident actions can cause harm, e.g., confusing ``disinfectant'' with ``hairspray'' (\figref{fig:intro}). Modern vision-language models (VLMs) are widely integrated into robot systems \citep{firoozi2025foundation}, yet remain prone to grounding errors, hallucinations, and overconfidence under out-of-distribution (OOD) objects, clutter, or ambiguous instructions. They typically operate as black boxes, with no accessible confidence. More fundamentally, any uncertainty derived from the VLM itself, whether raw logits, verbalized probability, or self-consistency, is a function of the same parameters $\theta_{\mathrm{vlm}}$ that produced the prediction. It therefore inherits the VLM's biases rather than providing an independent assessment \citep{zheng2024evaluating, padhi2025calibrating}. This raises a critical question: \textit{how can we derive reliable uncertainty for black-box VLMs that are decoupled from VLM-internal biases, to safely bound their behaviors in the open world?}

\pgap
To address this, we propose an \emph{external surrogate Bayesian model}. It formulates uncertainty estimation as an external statistical test on the VLM prediction, rather than a self-reported confidence drawn from $\theta_{\mathrm{vlm}}$. Assuming continuity of neural networks, the surrogate treats a held-out set $\mathcal{D}_{\mathrm{cal}}$ as empirical evidence. For each query, it estimates the posterior-predictive correctness within a local neighborhood via Beta-Bernoulli conjugacy, structurally decoupled from $\theta_{\mathrm{vlm}}$. The neighborhood is built by a hierarchical retrieval successively filtering in textual, cross-modal, and visual spaces, approximating a local neighborhood in the joint space without the curse of dimensionality. The surrogate yields two components. \textit{Aleatoric uncertainty} arises from local prediction-label inconsistency, indicating regions where the VLM \textit{systematically} fails. \textit{Epistemic uncertainty} arises from neighborhood sparsity, a principled signal for OOD queries in deterministic VLMs.

\pgap
Such uncertainty enables two active uncertainty alleviation extensions for visual grounding. \textit{Interactive visual grounding}~\citep{shridhar2018interactive, yang2022interactive} asks clarification questions to disambiguate the referring expression. We couple our per-object posterior with conformal prediction to form a $(1{-}\alpha)$-coverage prediction set, querying regions inside it until the target is identified. \textit{Active visual grounding}~\citep{xu2024towards, zhang2021invigorate} additionally takes physical actions, such as clearing obstacles or confirming targets, to actively gather evidence. We drive these decisions with a model-based policy that trades information gain against action cost. Under high uncertainty, both policies default to information-gathering rather than risky commitment.

\pgap
Experiments show consistent gains across state-of-the-art VLMs, in both \textit{discrimination} and \textit{calibration}. On a real Fetch robot, the system identifies OOD queries and improves robustness by resolving uncertainty before acting. Our core contribution is an \emph{external surrogate Bayesian model} that decouples uncertainty from the VLM itself, estimating closed-form aleatoric/epistemic components over a hierarchically retrieved neighborhood in $\mathcal{D}_{\mathrm{cal}}$. This, combined with conformal prediction and a model-based policy, realizes the robust uncertainty-to-action framework for active visual grounding in the open world.

\section{Related Work}

Foundation models increasingly enable general-purpose robotic autonomy, yet wrong decisions can cause high-stakes physical failures, making uncertainty estimation critical \cite{firoozi2025foundation}.

\pgap
\testit{Intrinsic Uncertainty for (V)LMs.} The most prevailing way for uncertainty estimation of VLMs is based on the model itself. \textit{Black-box} methods generate uncertainty in the outputs. For example, self-verbalized methods prompt the model to articulate its uncertainty \citep{kadavath2022language, linteaching}, yet are noisy and prompt-sensitive \citep{tian2023just, zhou2023navigating}. Consistency-based methods perturb inputs and measure semantic consistency \citep{khan2024consistency, kuhnsemantic}, which can be iterated to improve performance \citep{yadkori2024believe}. Yet, they are inefficient for real-time tasks and sometimes suffer from consistent errors \cite{padhi2025calibrating}. \textit{White-box} methods instead use internal representations, like token logits combined with conformal prediction \citep{ren2023robots, sun2024trustnavgpt, angelopoulos2021gentle}. Distance-aware methods can be further introduced for low-level failure detection \cite{xu2025can} and epistemic uncertainty modeling from latent-space geometry \citep{hashimoto2025efficient, venktesh2025sunar}, akin to kNN-based OOD detection \citep{sun2022out}. However, existing methods share a structural limitation. Black-box signals are a function of $\theta_{\mathrm{vlm}}$ and inherit its biases and miscalibration \cite{zheng2024evaluating}, while white-box methods require internal access and are inapplicable to state-of-the-art models.

\pgap
\testit{Uncertainty in Visual Grounding.} For grounding specifically, prior uncertainty estimation mainly relies on ensembles \citep{gal2017deep, miller2018dropout}, which require specific structures such as Bayesian neural networks. For black-box VLMs, recent work measures \textit{spatial consistency} of predicted boxes under perturbations \citep{zou2025uncertainty}. This is a localization variant of the consistency estimation above, sharing its efficiency limitations.

\pgap
\testit{Acting on Uncertainty.} Beyond uncertainty estimation, active visual grounding generates decisions of information-gathering for uncertainty alleviation. Classical formulations use POMDPs \cite{kaelbling1998planning} and belief-space planning \cite{kaelbling2013integrated}, recently extended to LLM-based interactive planning \citep{sun2024interactive}. For visual grounding, methods usually rely on human confirmation to clarify ambiguity before execution \citep{shridhar2018interactive, yang2022interactive, zhang2021invigorate}. We build on this line with our uncertainty estimator by closing the loop, where we estimate decoupled uncertainty externally and act on it within a unified decision policy.

\section{Problem Formulation}

\pgap
\paragraph{Uncertainty with Bayesian Surrogate}
\hanbo{We estimate per-query correctness of a black-box VLM.
We follow Bayesian formulation of uncertainties:
\begin{equation}
    p \triangleq \Pr(y \mid x, \mathcal{D}_{\mathrm{cal}}) = \mathbb{E}_{p(\theta \mid \mathcal{D}_{\mathrm{cal}})}\!\left[\Pr(y \mid x, \theta)\right].
    \label{eq:bayesian_posterior}
\end{equation}
where $\theta$ parameterize a model whose posterior is induced by a held-out labeled set $\mathcal{D}_{\mathrm{cal}}$.
Prior VLM-based estimators collapse this expectation onto the point estimate $\Pr(y \mid x, \theta_{\mathrm{vlm}})$, inheriting the VLM's biases.
Our surrogate instead maintains a distributional $\theta$ independent of $\theta_{\mathrm{vlm}}$, formalized via Beta-Bernoulli conjugacy on $\mathcal{D}_{\mathrm{cal}}$ (\secref{sec:quantification}).}

\pgap
\paragraph{Uncertainty of Visual Grounding} \hanbo{For visual grounding, a VLM is a mapping $\hat{f}: \ispc \times \tspc \to \tspc$ from an image $I \in \ispc$ and a free-form expression $\expr \in \tspc$ (e.g., ``pass me the nitro'') to regions $\rspc = \{\obj_i\}_{i=1}^{n}$ with binary predictions $\preds = \{\pred_i\}_{i=1}^{n}$.
Per region $\obj_i$, we instantiate \equationref{eq:bayesian_posterior} with query $x = (I, \expr, \obj_i, \pred_i)$ and label $y = \gt_i \in \{0,1\}$, where $\gt_i=1$ iff $\obj_i$ matches $\expr$. The held-out set is $\mathcal{D}_{\mathrm{cal}} = \{(I_j, \expr_j, r_j, y_j)\}_{j=1}^{N}$.
\tred{We aim to estimate the per-region posterior $p_{\obj}\!\triangleq\!\Pr(\gt_i\!=\!1 \mid I,\expr,\obj_i,\pred_i,\mathcal{D}_{\mathrm{cal}})$.}}
We include the detailed discussion for different models in \apref{sec:vg_models}.

\pgap
\paragraph{Active Visual Grounding}
We further extend this single-step formulation to steps $\step = 0, \dots, \mstep$, where the robot takes interactive actions $a_\step$ (asking, confirming, removing obstacles) to reduce uncertainty. At each step, the robot observes $(I_\step,\expr_\step)$, extracts regions $\rspc_\step$, and maintains a per-region belief $p_{r,\step}$ over all prior observations. The multi-step goal is to choose $a_\step$ that most reduces uncertainty and to return $\rspc^{*}\subseteq\rspc_\step$ containing all specified targets (\secref{sec:policy}).

\section{Uncertainty Estimation for Visual Grounding}
\label{sec:uncertainty}

\begin{figure}[t]
    \centering
    \includegraphics[width=\textwidth]{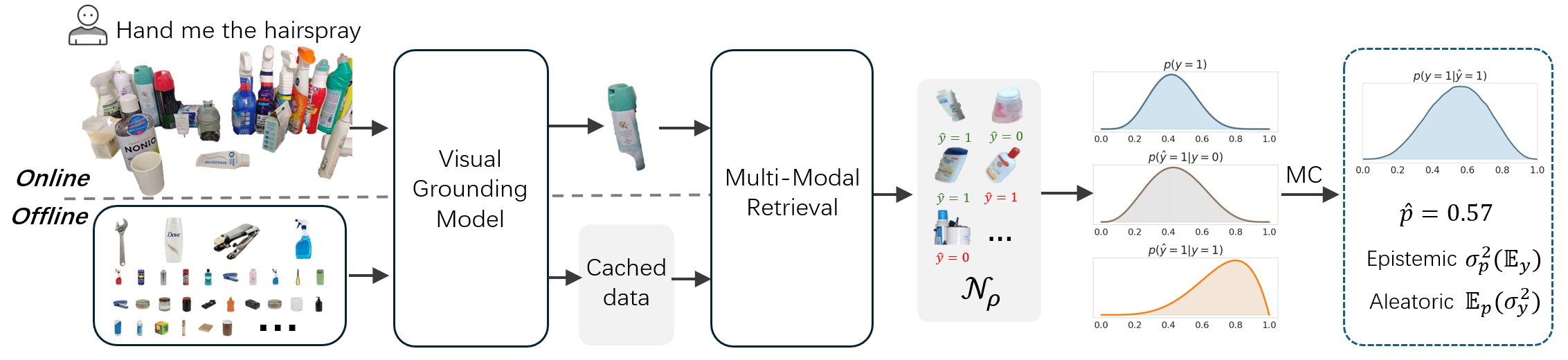}
    \caption{Overview of the visual grounding uncertainty estimation based on neighborhood empirical correctness.}
    \label{fig:overview}
    \vspace{-0.6cm}
\end{figure}

\pgap
As illustrated in \figref{fig:overview}, our approach comprises two components: an uncertainty quantification method over a local neighborhood of $\mathcal{D}_{\mathrm{cal}}$ followed by a calibrator aligning predicted probabilities with empirical correctness (\secref{sec:quantification}), and a hierarchical retrieval that constructs query-specific neighborhoods in multi-modal high-dimensional spaces (\secref{sec:retrieval}).

\subsection{Uncertainty Quantification}
\label{sec:quantification}

\pgap
We formulate uncertainty quantification through a Bayesian framework that factorizes the correctness probability into a data prior and a model likelihood, estimated from retrieved evidence in a local neighborhood of $\mathcal{D}_{\mathrm{cal}}$ to ensure robustness under data imbalance.

\pgap
\paragraph{Retrieval-based Evidence}
Let $h_j = \phi(I_j, \expr_j, r_j)$ be the offline representation of samples in $\mathcal{D}_{\mathrm{cal}}$. For a test query with representation $h = \phi(I,\expr,\obj)$, we form a local neighborhood $\mathcal{N}(h) \subset \mathcal{D}_{\mathrm{cal}}$ of samples close to $h$ under a distance metric $d(\cdot, \cdot)$.
Prediction uncertainty for region $\obj$ and prediction $\pred$ then arises from two sources: (i) insufficient evidence in $\mathcal{N}(h)$ (epistemic); (ii) prediction-label inconsistency within $\mathcal{N}(h)$ (aleatoric).

\pgap
\paragraph{Probabilistic Modeling} We disentangle $\Pr(y \mid \pred, \mathcal{N})$ into a data-distributional prior and a model-predictive likelihood:
\begin{equation}
\hat{p}=\Pr(y \mid \pred, \mathcal{N}) \propto {\Pr(\pred \mid y, \mathcal{N})} \cdot {\Pr(y \mid \mathcal{N})}. \nonumber
\end{equation}
This factorization is crucial under data imbalance, which otherwise undermines model-prediction reliability. Since both $y$ and $\pred$ are binary, we model them as Bernoulli distributions with Beta priors, exploiting Beta-Bernoulli conjugacy for analytical parameter distributions given evidence in $\mathcal{N}$:
\begin{align}
y \mid \mathcal{N} &\sim \mathrm{Bern}(\theta_y), &\theta_y  &\sim \mathrm{Beta}(\alpha_y, \beta_y),
\nonumber \\
\pred \mid y, \mathcal{N}
&\sim \mathrm{Bern}(\theta_{\pred\mid y}),
&\theta_{\pred\mid y}
&\sim \mathrm{Beta}(\alpha_{\pred\mid y}, \beta_{\pred\mid y}),
\label{eq:bernbeta}
\end{align}
where $\theta_y, \theta_{\pred|y}$ are the latent parameters of the prior and likelihood, with hyperparameters $(\alpha, \beta)$ updated in closed form from softened counts over $\mathcal{N}(h)$ that down-weight farther neighbors (\apref{ap:kernel_details}).
Based on $\hat{p}$, we further calibrate on $\mathcal{D}_{\mathrm{cal}}$ to obtain $\tilde{p}$ satisfying $\Pr(y=1\mid \tilde{p}=p)=p$ via standard post-hoc methods \citep{guo2017calibration} (\apref{ap:calibration}).

\pgap
\paragraph{Uncertainty Decomposition}
We adopt the variance-based decomposition of predictive uncertainty \cite{depeweg2018decomposition}, which partitions total uncertainty into \textit{aleatoric} and \textit{epistemic} components following the \textit{law of total variance}:
\begin{equation}
    \sigma^2(y)
    =
    \underbrace{
        \sigma^2_{p(\theta)}\!\big(
            \mathbb{E}[y|\theta]
        \big)
    }_{\text{epistemic}}
    +
    \underbrace{
        \mathbb{E}_{p(\theta)}\!\big[
            \sigma^2(y|\theta)
        \big]
    }_{\text{aleatoric}}.
    \label{eq:total_var}
\end{equation}
The product $\Pr(y \mid \pred, \mathcal{N})$ of the Beta factors (Eq.~\ref{eq:bernbeta}) is not analytical, so we estimate Eq.~\ref{eq:total_var} by drawing $S$ samples from the Beta posteriors and computing realizations $\hat{p}^{(s)}$:
\begin{align}
u^{\mathrm{epi}}
\approx \frac{1}{S-1} \sum_{s=1}^{S}
\left(\hat{p}^{(s)} - \bar{p}\right)^2,\:
u^{\mathrm{ale}}
\approx \frac{1}{S} \sum_{s=1}^{S}
\hat{p}^{(s)}\!\left(1 - \hat{p}^{(s)}\right), \nonumber
\end{align}
with $\bar{p}$ the empirical mean.
Intuitively, $u^{\mathrm{epi}}$ captures lack of evidence, which means a sparse $\mathcal{N}$ induces high variance in the sampled $\theta$. $u^{\mathrm{ale}}$ captures data inconsistency, which means high Bernoulli variance even under a sharp Beta posterior.

\subsection{Neighborhood Retrieval}
\label{sec:retrieval}

\pgap
Since the high-dimensional multi-modal space is sparse with respect to $\mathcal{D}_{\mathrm{cal}}$, we construct $\mathcal{N}(h)$ via a hierarchical retrieval with progressively refined filtering. Given a query $(I,\expr,\obj)$ with representation $(h^v,h^\expr)=\phi(I,\expr,\obj)$, we apply three similarity filters in turn: (i) text similarity over $\phi_\expr(\expr_j)$ via a sentence-transformer \citep{reimers2019sentence}; (ii) cross-modal distance via CLIP \citep{radford2021learning}; (iii) visual distance over $\phi_v(I_j)$ via DINOv2 \citep{oquab2024dinov2}. Each stage produces a similarity score that softly down-weights farther neighbors, with details discussed in \apref{ap:kernel_details}.
The composed weights define the local neighborhood $\mathcal{N}(h)$ in the joint representation space.
It avoids the curse of dimensionality of single-stage high-dimensional retrieval.
Our retrieval can generalize standard held-out usage \cite{guo2017calibration}, which corresponds to globally uniform weights over $\mathcal{D}_{\mathrm{cal}}$.
Hence, the retrieval governs the trade-off between data coverage and locality in a unified framework.

\section{Active Visual Grounding}

\begin{wrapfigure}{r}{0.57\textwidth}
    \centering
    \includegraphics[width=0.55\textwidth]{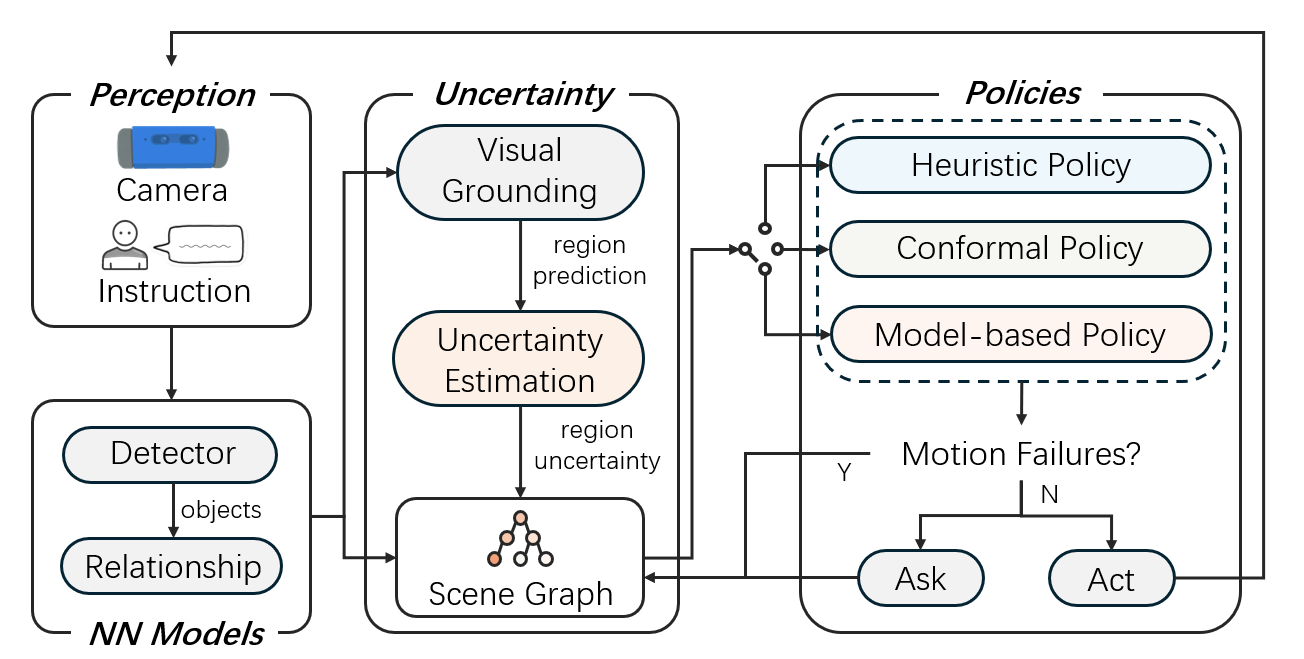}
    \caption{Overview of active visual grounding.}
    \label{fig:interact_sys}
    \vspace{-0.5cm}
\end{wrapfigure}

\pgap
We formulate uncertainty-aware visual grounding in the real world as an active visual grounding problem \cite{shridhar2018interactive, zhang2021invigorate}, where the robot identifies target objects under open-ended instructions while actively reducing grounding uncertainty through interaction and physical actions (\figref{fig:interact_sys}).
Building on reactively estimated uncertainty for black-box visual grounding, we design interactive policies that reason about future uncertainties over multiple steps (\secref{sec:policy}), together with uncertainty tracking over the tracked regions $\{\mathcal{R}_\step\}_{\step=0}^{\mstep}$ (\secref{sec:belief_tracking}).

\subsection{Interactive Policies}
\label{sec:policy}

\pgap
At each step $\step$, the robot maintains object uncertainty $p_{\obj,\step},\:{r\in\mathcal{R}_\step}$ from all historical observations $\{(I_{\step'},\expr_{\step'}, \mathcal{R}_{\step'})\}_{\step'=0}^{\step}$ and chooses an action $a_\step\in\act$ that most efficiently reduces uncertainty over $\mathcal{R}_\step$, where $\act$ comprises asking an open question ($\acti$), confirming an object ($\actc$), grasping a target ($\actg$), and removing an obstacle ($\actr$).
We use two policies, with full formulations given in \apref{ap:policy}.

\pgap
\paragraph{Model-based Policy}
The model-based policy searches over action sequences that trade information gain against action and interaction cost, evaluating actions by their expected effect on the uncertainty distribution $\{p_{\obj,\step}\}$: $\pi^*= \arg\max_{\pi}\;\mathbb{E}\!\left[\sum_{\step=0}^{T} r(s_\step, a_\step)\right]$, where $s_\step$ is the state at step $\step$. The reward $r(s_\step,a_\step)$ trades task completion against interaction cost. The factorized object-level state, observation, transition models, and reward functions used during forward search are detailed in \apref{ap:planning}.

\pgap
\paragraph{Conformal-prediction Policy}
Following \cite{ren2023robots}, this policy applies conformal prediction on $p_{\obj,\step}$ to form a prediction set $\mathcal{C}_\step = \{ r \in \mathcal{R}_\step \mid p_{\obj,\step} \ge \tau(\alpha) \}\subseteq\rspc_\step$ that covers the ground-truth target with probability $1-\alpha$, where the nonconformity threshold $\tau(\alpha)$ is calibrated on $\mathcal{D}_{\mathrm{cal}}$ at a manually selected risk level $\alpha$. The robot then confirms regions in $\mathcal{C}_\step$ until the target is identified.
We further use an \emph{object-wise} variant that tests each region independently and acts greedily, grasping or removing when unambiguous and otherwise asking, with full details in \apref{ap:conformal}.

\pgap
\paragraph{Hierarchical Planning}
A hierarchical planner additionally filters actions by physical feasibility, approximated via visual relationship priors \cite{zhang2018visual} that prune unreachable or blocked targets. On planning or action failures, the system updates the constraint model and replans (\apref{ap:hierarchy}).

\subsection{Sequential Uncertainty Tracking}
\label{sec:belief_tracking}

\pgap
We track object-centric uncertainty sequentially after each interaction.
At step $\step$, $p_{\obj,\step}=p(y=1\mid I_t,\expr_{0:t},\obj)$ is the probability that region $\obj$ is the target conditioned on the visual observation $I_t$ and all language observations $\expr_{0:\step}$.
Assuming conditional independence of grounding across expressions, it decomposes recursively (derivation in \apref{ap:derv_uncert_track}) as
\begin{equation}
    p_{\obj,\step}\propto \eta\:\frac{p(y\mid I_t,\expr_t,\obj)}{p(y\mid I_t,\obj)}\:p_{\obj,\step-1},
\end{equation}
where $p(y\mid I_t,\obj)$ and $p(y\mid I_t,\expr_t,\obj)$ are the prior and the single-step estimate, and $\eta$ is a normalization constant.
For repeated visual frames after grasping or removing actions, na\"{i}ve Bayesian updates overcount highly correlated views and yield overconfident estimates. We instead apply a moving average $p' = \omega p + (1-\omega)p_{\text{new}}$, where $\omega$ controls the memory horizon. This follows linear opinion pooling~\cite{genest1986combining}, which drops the independence assumption violated by correlated views~\cite{thrun2002probabilistic}.
Newly detected objects are initialized with a uniform prior.

\section{Experiments}

\pgap
Our experiments evaluate three aspects of the proposed approach: uncertainty quality across VLMs and benchmarks (\secref{sec:bench_exp}); interactive visual grounding with asking policies (\secref{sec:interaction_exp}); and active visual grounding on a real robot (\secref{sec:robot_exp}).

\subsection{Benchmarking Uncertainty Quality}
\label{sec:bench_exp}

\pgap
We evaluate on the RefCOCO series~\citep{yu2016modeling} and RoboRefIt~\citep{lu2023vl}, applying our estimator to three VLM grounding paradigms: pointing (Molmo~\citep{deitke2025molmo}), detection (Qwen2.5VL-72B-AWQ~\citep{bai2025qwen2, lin2024awq}), and region scoring (Qwen2.5VL-SoM~\citep{yang2023set}). We report Recall@1 (IoU$>$0.5), AUPRC, and ECE/aECE/NLL/Brier. Full metric definitions are in \apref{ap:cali_metrics}, with extended per-metric results in \apref{ap:bench_result}.

\paragraph{How does our method perform against existing baselines?}

\begin{table}[t]
\centering
\caption{Comparison across grounding models and test sets (Recall@1~/~AUPRC).}
\label{tab:summary}
\resizebox{\textwidth}{!}{%
\begin{tabularx}{1.15\textwidth}{
>{\raggedright\arraybackslash}p{2.8cm}l
*{2}{>{\centering\arraybackslash}X}  %
*{2}{>{\centering\arraybackslash}X}  %
>{\centering\arraybackslash}X        %
*{2}{>{\centering\arraybackslash}X}  %
}
\toprule
\multirow{2}{*}{\textbf{Model}} & \multirow{2}{*}{\textbf{Method}} &
\multicolumn{2}{c}{\textbf{RefCOCO}} &
\multicolumn{2}{c}{\textbf{RefCOCO+}} &
\textbf{RefCOCOg} &
\multicolumn{2}{c}{\textbf{RoboRefIt}} \\
\cmidrule(lr){3-4}\cmidrule(lr){5-6}\cmidrule(lr){7-7}\cmidrule(lr){8-9}
& & \textbf{testA} & \textbf{testB} & \textbf{testA} & \textbf{testB} & \textbf{test} & \textbf{testA} & \textbf{testB} \\
\midrule
\multirow{2}{*}{\makecell[l]{Molmo\\({\it Pointing})}}
& Base  & 78.3/69.5 & 67.7/41.5 & 76.2/64.2 & 60.0/32.6 & 66.1/46.8 & 66.9/44.4 & 66.9/49.4 \\
& Ours  & \textbf{79.6}/\textbf{80.9} & \textbf{70.2}/\textbf{57.2} & \textbf{77.6}/\textbf{77.8} & \textbf{62.3}/\textbf{48.9} & \textbf{69.5}/\textbf{59.6} & \textbf{69.1}/\textbf{66.6} & \textbf{71.3}/\textbf{67.1} \\
\midrule
\multirow{2}{*}{\makecell[l]{Qwen2.5VL$^\dagger$\\({\it Detection})}}
& Base  & 82.6/80.0 & 77.3/62.2 & 78.4/70.5 & 67.1/47.1 & 77.2/66.6 & 76.4/60.3 & 75.9/62.3 \\
& Ours  & \textbf{83.2}/\textbf{86.0} & \textbf{78.6}/\textbf{71.6} & \textbf{79.2}/\textbf{79.1} & \textbf{69.1}/\textbf{55.7} & \textbf{77.7}/\textbf{72.8} & \textbf{77.7}/\textbf{77.1} & \textbf{77.3}/\textbf{73.4} \\
\midrule
\multirow{2}{*}{\makecell[l]{Qwen2.5VL$^\dagger$+SoM\\({\it Scoring})}}
& Base  & 64.3/43.5 & 50.6/22.9 & 61.3/38.6 & 43.5/17.8 & 54.7/29.8 & 46.3/23.9 & 41.1/20.1 \\
& Ours  & \textbf{65.1}/\textbf{66.5} & \textbf{52.1}/\textbf{40.0} & \textbf{62.5}/\textbf{62.1} & \textbf{45.1}/\textbf{32.8} & \textbf{56.7}/\textbf{46.8} & \textbf{50.8}/\textbf{47.3} & \textbf{46.8}/\textbf{41.5} \\
\bottomrule
\multicolumn{9}{l}{$^\dagger$ Qwen2.5VL is the 72B AWQ-quantized~\citep{lin2024awq} version due to computational resource limitation.}
\end{tabularx}%
}
\end{table}

\begin{figure}[t]
    \centering
    \vspace{-0.3cm}
    \includegraphics[width=0.85\textwidth]{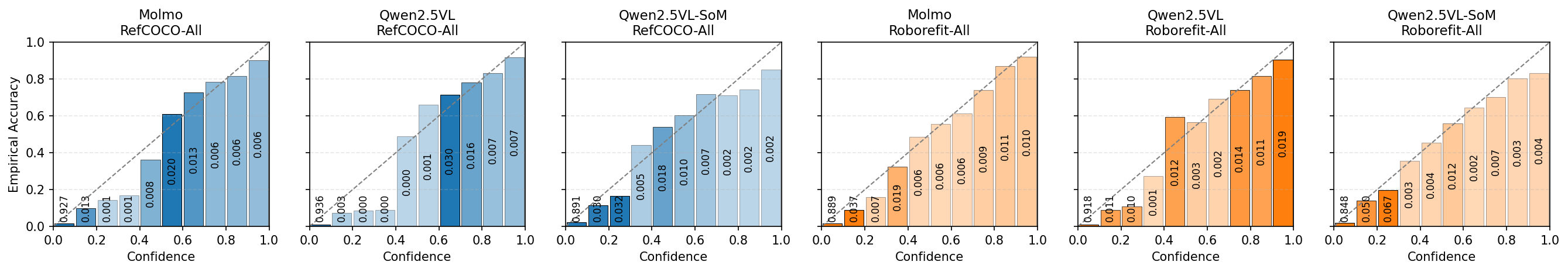}
    \caption{Reliability curves of our method, comparing predicted probability against empirical correctness across VLMs and datasets. Detailed reliability diagrams for all models and baselines are in \apref{ap:bench_result}.}
    \label{fig:reliability_curves}
    \vspace{-0.5cm}
\end{figure}

\begin{wraptable}{r}{0.6\textwidth}
\vspace{-0.6cm}
\centering
\caption{Consistency-based ablation on RefCOCO (R@1/AUPRC).}
\label{tab:consistency_ablation}
\resizebox{0.6\textwidth}{!}{%
\begin{tabularx}{0.69\textwidth}{l *{4}{>{\centering\arraybackslash}X}}
\toprule
\multirow{2}{*}{\textbf{Method}} & \multicolumn{2}{c}{\textbf{Molmo} ({\it Point})} & \multicolumn{2}{c}{\textbf{Qwen2.5VL}$^\dagger$ ({\it Det})} \\
\cmidrule(lr){2-3}\cmidrule(lr){4-5}
 & \textbf{testA} & \textbf{testB} & \textbf{testA} & \textbf{testB} \\
\midrule
Base       & 78.3/69.5 & 67.7/41.5 & 82.6/80.0 & 77.3/62.2 \\
C-$N_3$    & 79.6/78.8 & 67.6/55.7 & 83.5/87.9 & 79.4/74.6 \\
C-$N_5$    & 81.4/82.5 & 68.9/59.7 & 83.8/89.9 & 78.4/76.6 \\
Ours       & 79.6/80.9 & 70.2/57.2 & 83.2/86.0 & 78.6/71.6 \\
Ours-$N_3$ & 81.0/86.4 & 71.2/67.7 & 84.3/91.6 & 80.0/80.4 \\
Ours-$N_5$ & \textbf{82.6}/\textbf{88.3} & \textbf{72.8}/\textbf{70.4} & \textbf{84.6}/\textbf{92.3} & \textbf{80.1}/\textbf{81.8} \\
\bottomrule
\multicolumn{5}{l}{\footnotesize $^\dagger$ Qwen2.5VL is the 72B AWQ-quantized~\citep{lin2024awq} version.}
\end{tabularx}%
}
\vspace{-0.4cm}
\end{wraptable}
We compare \textit{Ours} against the no-uncertainty \textit{Base} (\tabref{tab:summary}) and the consistency-based \textit{C-$N_K$}~\citep{khan2024consistency, padhi2025calibrating} (\tabref{tab:consistency_ablation}), which samples the VLM on $K$ perturbed copies; \textit{Ours-$N_K$} fuses our surrogate with the same sampling.
Across all grounding models and splits, \textit{Ours} consistently improves Recall@1 and AUPRC, showing that the estimated uncertainty reliably ranks correct against incorrect predictions while improving the accuracy of visual grounding tasks.
Against the consistency baseline, \textit{Ours} is consistently better at controlled inference cost.
Besides, by results of \textit{Ours-$N_K$}, it demonstrates that our method is scalable to get better performance during inference time, similar to the consistency-based methods.
Finally, reliability diagrams (\figref{fig:reliability_curves}) further confirm that predicted probabilities can track empirical correctness accurately, reflecting the underlying true probabilities. More detailed comparisons and discussions are provided in \apref{ap:bench_result}.

\paragraph{How sensitive is it to the embedding space?}
\begin{wraptable}{r}{0.5\textwidth}
\vspace{-0.8cm}
\centering
\caption{Retrieval-embedding ablation on RefCOCO with Molmo (R@1/AUPRC).}
\label{tab:retrieval_ablation}
\resizebox{0.5\textwidth}{!}{%
\begin{tabularx}{0.575\textwidth}{l *{4}{>{\centering\arraybackslash}X}}
\toprule
{\textbf{Split}} & {\textbf{Base}} & {\textbf{Text-only}} & {\textbf{CLIP-only}} & {\textbf{Ours}} \\
\midrule
testA & 78.3/69.5 & 78.3/69.6 & 78.3/75.2 & {\textbf{79.6}}/{\textbf{80.9}} \\
testB & 67.7/41.5 & 67.7/41.5 & 69.1/55.9 & {\textbf{70.2}}/{\textbf{57.2}} \\
\bottomrule
\end{tabularx}%
}
\vspace{-0.4cm}
\end{wraptable}
Our retrieval uses three off-the-shelf external embeddings (Sentence-BERT~\citep{reimers2019sentence}, CLIP~\citep{radford2021learning}, DINOv2~\citep{oquab2024dinov2}) covering text, cross-modal, and visual similarity. Though they are not trained to align with the VLM's decision boundary, results show that our surrogate model consistently benefits from retrieving using all these models.
From \tabref{tab:retrieval_ablation}, we can see that even a single CLIP cross-modal filter alone boosts both the performance of accuracy and ranking consistently, and the full three-stage hierarchy brings more gains.

\paragraph{How robust is it against domain shift?}
\begin{wrapfigure}{r}{0.5\textwidth}
    \centering
    \includegraphics[width=0.48\textwidth]{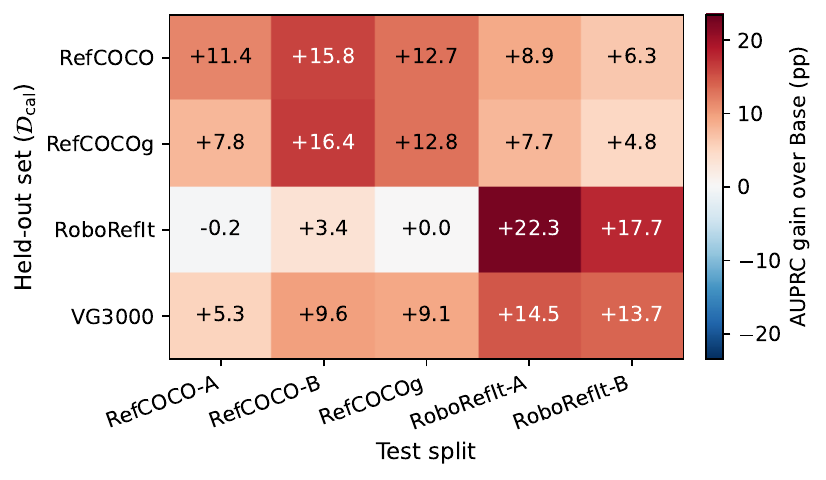}
    \vspace{-0.4cm}
    \caption{AUPRC gain of \textit{Ours} over the no-uncertainty \textit{Base} on Molmo, across held-out (rows) and test (columns) splits.}
    \label{fig:cross_dataset_gain}
    \vspace{-0.4cm}
\end{wrapfigure}
We have done experiments across domains to test robustness of our method against domain shifts.
Across all held-out and test pairs on Molmo (\figref{fig:cross_dataset_gain}), in-domain pairs gain up significantly while out-of-domain pairs (e.g., using RoboRefIt as the held-out set for RefCOCO) do not suffer from performance drop. It shows a safe floor under drift, which is expected since OOD will be always triggered, and the uncertainty falls back to bias towards the  model prediction due to lack of priors. Particularly, we include VG3000 as a more diverse and in-the-wild representative held-out dataset, which is a subset from Visual Genome \cite{krishna2017visual}.  Its uniformly positive gains show that diverse public datasets can be leveraged when in-domain collection is infeasible. Together, these results confirm that our uncertainty estimation is robust across diverse domain shifts.
To test robustness against noise, we further add controlled input perturbations and test the performance, as shown in \apref{ap:input_ablation}. Additional qualitative examples, calibration ablations, and inference-cost analyses are provided in \apref{ap:bench_result}.

\subsection{Interactive Visual Grounding}
\label{sec:interaction_exp}

\begin{wrapfigure}{r}{0.5\textwidth}
    \vspace{-1.3cm}
    \centering
    \includegraphics[width=0.48\textwidth]{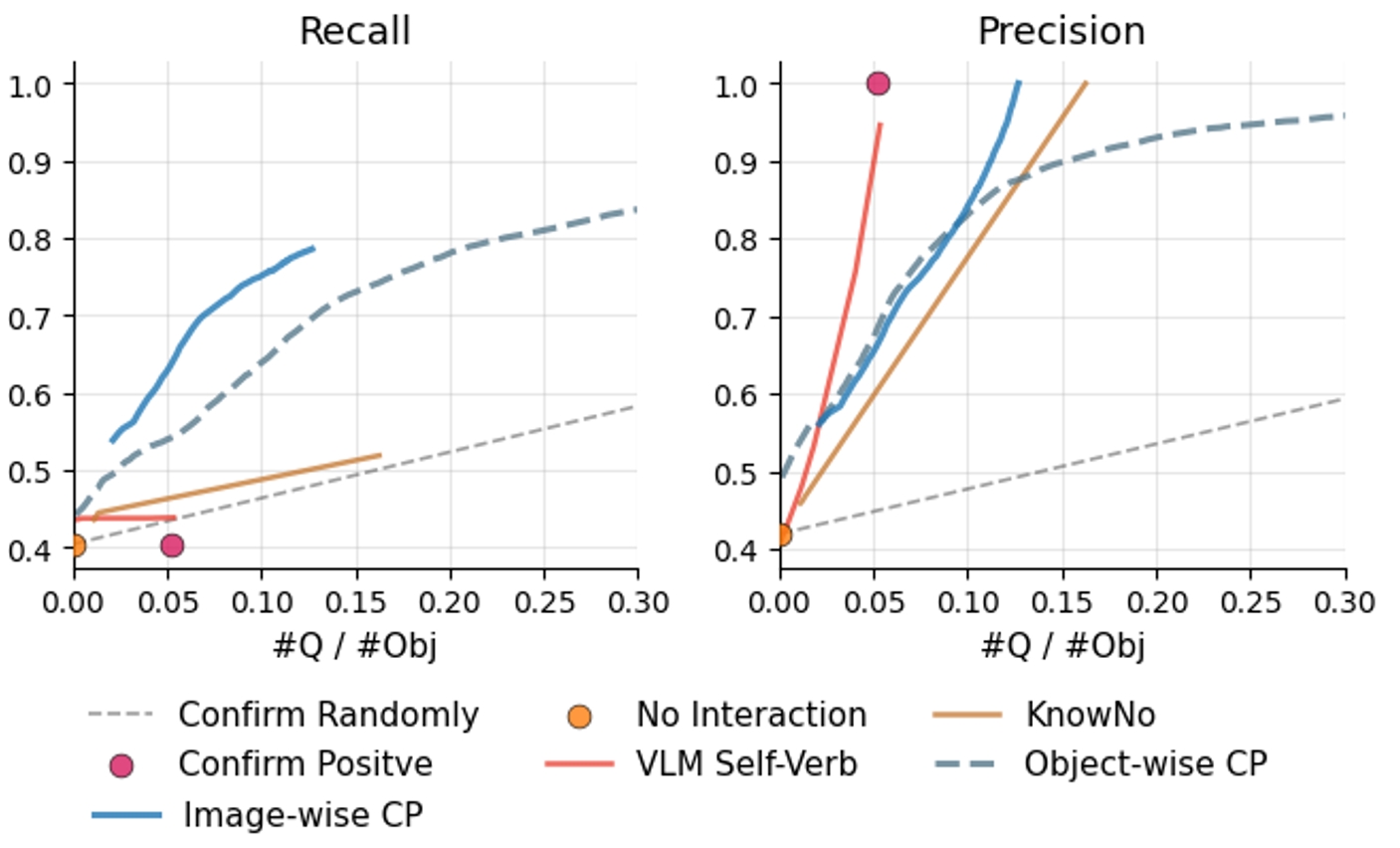}
    \vspace{-0.2cm}
    \caption{IVG results on RoboRefIt: recall (left) and precision (right) vs.\ question budget.}
    \label{fig:ivg_results}
    \vspace{-0.5cm}
\end{wrapfigure}
We evaluate how estimated uncertainty enables effective information gathering in interactive visual grounding~\citep{shridhar2018interactive}, where robots may ask questions to ground targets.
Experiments on RoboRefIt measure task accuracy against the simulated number of confirmations, isolating the impact of uncertainty under minimal interaction cost.
Without uncertainty, baselines include two trivial policies (\textit{Confirm Randomly}, \textit{Confirm Every}) and \textit{Confirm Positive}~\citep{shridhar2018interactive, zhang2021invigorate}. With uncertainty, we include \textit{VLM Self-Verbalization}~\citep{linteaching} and \textit{KnowNo}~\citep{ren2023robots}. Ours integrates conformal prediction in \textit{Image-wise} and \textit{Object-wise} variants (\secref{sec:policy}; details in \apref{ap:ivg_baselines}) with physical actions disabled. All runs use Qwen2.5VL-SoM, since Molmo does not generate consistent per-object scores.
Both \textit{Object-wise} and \textit{Image-wise CP} consistently perform better than all baselines (\figref{fig:ivg_results}), with higher accuracy and recall at every question budget.
\textit{Image-wise CP} reaches high performance rapidly under low budgets via joint image-level reasoning, but is limited by top-$5$ truncation at higher budgets. \textit{Object-wise CP} avoids truncation by treating each object independently, at slight efficiency cost. Further details can be found in \apref{ap:ivg_result_details}.

\subsection{Active Visual Grounding on a Real Robot}
\label{sec:robot_exp}

\begin{wrapfigure}{r}{0.5\textwidth}
    \vspace{-1.7cm}
    \centering
    \includegraphics[width=0.48\textwidth]{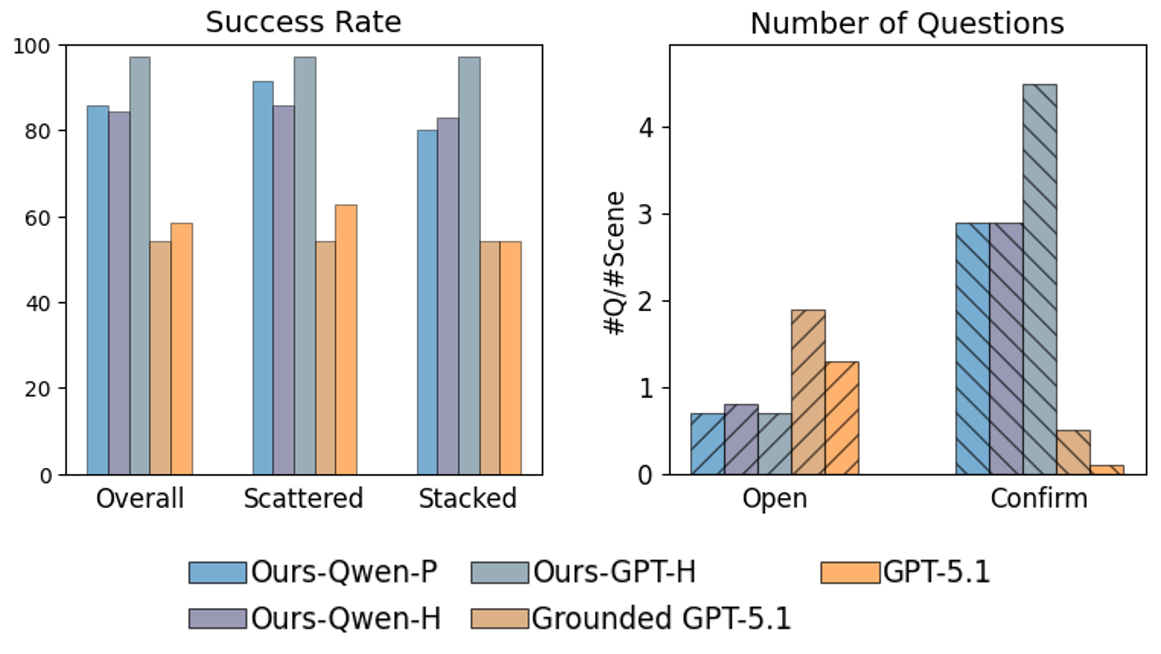}
    \vspace{-0.2cm}
    \caption{Real-world experimental results on a Fetch robot.}
    \label{fig:real_robot}
    \vspace{-0.4cm}
\end{wrapfigure}
We validate on a real Fetch robot with the uncertainty estimator trained only on joint public datasets, without in-domain data collection. The robot follows open-world instructions and may actively remove obstacles and ask questions to resolve uncertainty.
Instructions are deliberately multi-target and category-level, e.g., \textit{``pick all medical objects''}, so manipulation continues after the first retrieval until no plausible target remains rather than terminating on a single grasp.
We test \textit{Scattered} (isolated objects) and \textit{Stacked} (bin-picking~\citep{mahler2017learning} with heavy clutter) scenes (5 scenes each, 7 objects/scene, 70 trials in total), reporting \textit{Success Rate} and \textit{\#Q/\#Scene}.
We implement two heuristic policies (\textit{Ours-Qwen-H}, \textit{Ours-GPT-H}) and a model-based policy (\textit{Ours-Qwen-P}). These are compared against \textit{GPT-5.1}, an end-to-end LLM policy without uncertainty modeling~\citep{tziafas2024towards}, and \textit{Grounded GPT-5.1}, which runs GPT-5.1 decisions on our perception pipeline but ablates explicit uncertainty reasoning (\apref{ap:robot_baselines}).

\begin{figure}[t]
    \centering
    \includegraphics[width=\textwidth]{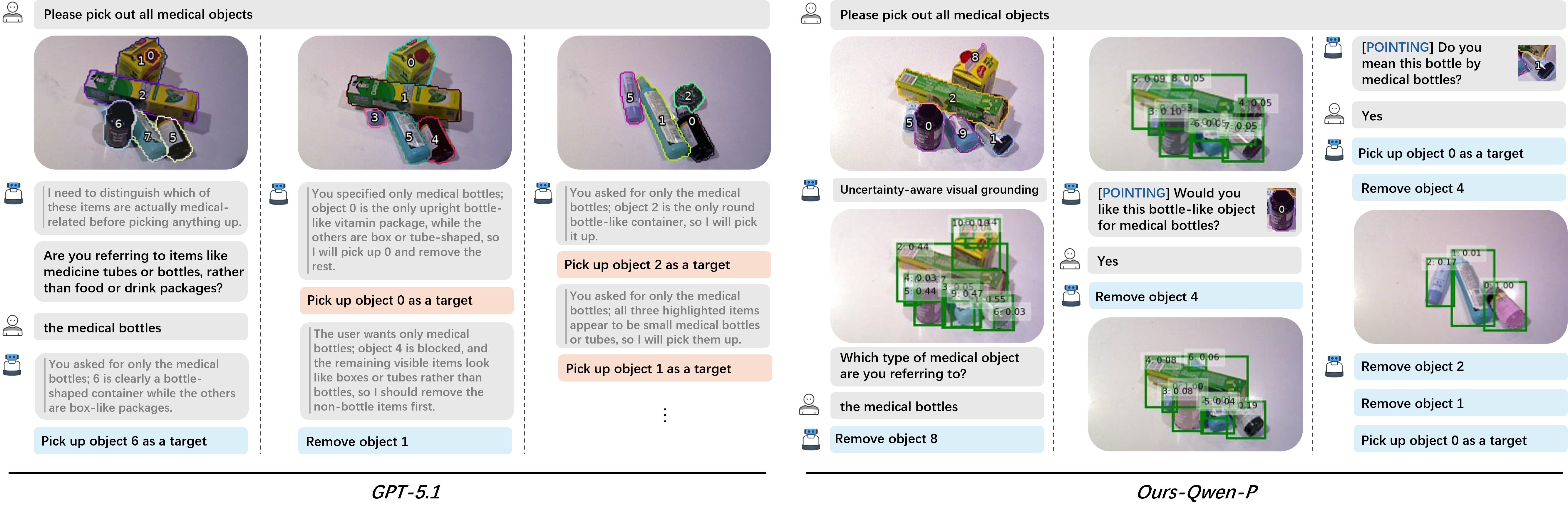}
    \caption{Qualitative comparison of real-world experiments between GPT-5.1 as the policy and our uncertainty-aware policy. Left: a GPT-5.1 roll-out; Right: a roll-out of our method (Ours-Qwen-P).}
    \label{fig:real_robot_qual}
    \vspace{-0.6cm}
\end{figure}

\pgap
Across Scattered and Stacked scenes, uncertainty-aware variants significantly outperform LLM-only baselines ($p<0.001$, $\chi^2$-test, 70 samples; \figref{fig:real_robot}). The gains come from explicit uncertainty modeling, which prioritizes removing obstacles that most likely occlude targets and asks when uncertain, whereas LLMs make overconfident commitments to incorrect targets.
Notably, \textit{Grounded GPT-5.1} does not improve over \textit{GPT-5.1} despite using our perception pipeline. This isolates premature action commitment, not perception, as the dominant LLM failure mode. As shown in \figref{fig:real_robot_qual}, GPT-5.1 relies on implicit instruction assumptions, a form of confirmation bias~\citep{klayman1995varieties, wan2025unveiling}; our method instead acts only after interaction gathers sufficient information.
A per-mode failure breakdown (\apref{ap:fail_analysis}) shows grounding errors (GRD) dominate for the LLM baselines, while detection, grasping, and relation failures are rare ($\leq 2$). Thus the gap is attributable to overconfident grounding rather than perception or manipulation.

\section{Conclusion}

\pgap
\paragraph{Conclusion}
We present a principled uncertainty estimation framework for black-box vision-language models, grounded in empirical correctness over a hierarchically retrieved local neighborhood of a held-out calibration set. By decomposing uncertainty into aleatoric and epistemic components via Beta-Bernoulli conjugacy, the surrogate yields well-calibrated and interpretable estimates that are structurally decoupled from the VLM's internal biases. Coupled with conformal prediction and a model-based policy, it turns these estimates into robust, uncertainty-aware actions for active visual grounding. Across grounding benchmarks and state-of-the-art VLMs, it consistently improves both discrimination and calibration. On a real Fetch robot, it enables principled trade-offs between confident execution and information gathering, substantially improving robustness under open-world instructions.

\pgap
\paragraph{Limitations and Future Work}
Our approach relies on object-centric representations that assume object independence, which may not hold in cluttered or relational scenes. In the future, structured scene graphs that jointly model objects and relations are a natural extension for context-aware uncertainty estimation. The neighborhood retrieval also depends on a representative held-out set, which further motivates data-scalable open-world uncertainty estimation integrated with learning-based retrieval to broaden coverage at lower cost. Besides, an uncertainty-aware manifold may be preferable to the pretrained embedding models. It is also valuable to extend our uncertainty estimator for object-goal navigation and active-viewpoint settings, where uncertainty is crucial to improve robustness and efficiency. Finally, integrating the semantic and symbolic uncertainty with long-horizon, uncertainty-aware task and motion planning is promising for future work, particularly for high-stakes tasks.

\bibliography{refs}

\clearpage
\appendix
\section{Appendix}

\subsection{Uncertainty-aware Visual Grounding}
\label{sec:vg_models}

Different VLM models produce visual grounding predictions in diverse forms, which can generally be categorized into three types: pointing models, detection models, and scoring models.
To unify these heterogeneous architectures under our formulation, we assume a high-recall open-set detector (e.g., OWL-ViT \citep{minderer2023scaling}) is used to propose an inclusive set of candidate regions $\mathcal{R}(I)$, ensuring that potential targets are included.
Given an image $I$ and a referring expression $\expr$, each candidate region $\obj\in\mathcal{R}(I)$ is associated with a binary label indicating whether it matches the referred object.
We will elaborate on how this assumption is achieved in the following.

\subsubsection{Pointing Models}
Point-based models, such as Molmo \citep{deitke2025molmo}, predict one or multiple image coordinates $(x,y)$ indicating the most relevant point(s) for a given query.
We associate predicted points with detected regions $\mathcal{R}(I)$ via Hungarian matching on the Euclidean distance from each point to the region centroid.
A region $\obj$ receives binary prediction $\hat{y}=1$ if it is matched with a predicted point and $\hat{y}=0$ otherwise.
This mapping adapts point-based models to the region-query form.

\subsubsection{Detection Models}
End-to-end detection-based grounding models such as Qwen2.5-VL \citep{bai2025qwen2} directly output bounding boxes conditioned on the query, which we treat as positive predictions $\hat{y}=1$.
To obtain negative predictions, we match each predicted box against proposals in $\mathcal{R}(I)$ by Intersection-over-Union (IoU).
Proposals with $\mathrm{IoU} \ge 0.5$ to any predicted box are kept as positives and the rest are marked $\hat{y}=0$.
We choose $\mathrm{IoU} \ge 0.5$ to match the standard detection-quality criterion used by Recall@1 in \secref{sec:bench_exp}, so the binary prediction is consistent with the evaluation metric.

\subsubsection{Scoring Models}
Classical scoring grounding models, such as INGRESS \citep{shridhar2018interactive}, and recent set-of-mask prompting \citep{yang2023set} inherently follow our formulation.
They operate on a predefined region set $\mathcal{R}(I)$ and output region-query matching scores that can be thresholded to produce binary predictions $\hat{y}\in\{0,1\}$.
No additional processing is required.

\subsection{Details of Weighted Neighborhoods}
\label{ap:kernel_details}

The neighborhood $\mathcal{N}(h)$ is implemented via per-sample soft weights rather than a hard radius, so a sample's contribution decreases continuously with distance and the method has no sensitive cutoff hyperparameter.
Each of the three stages produces a soft weight via an exponential decay
\begin{equation}
w_j^{(m)} \;=\; \exp\!\bigl(-\gamma\, d_m(h,\, h_j)\bigr), \quad m \in \{\text{text},\,\text{cross},\,\text{visual}\}, \nonumber
\end{equation}
where $d_m$ is the cosine distance in the corresponding embedding space and $\gamma = 10$ is a fixed bandwidth shared by all three stages.
We adopt this exponential-decay form because it smoothly down-weights distant neighbors without a hard cutoff and is numerically stable over the bounded cosine-distance range.
The composed per-sample weight is $w_j \;=\; w_j^{(\text{text})} \cdot w_j^{(\text{cross})} \cdot w_j^{(\text{visual})}$, combining the three stages multiplicatively so a sample contributes only when all three agree.

The Beta hyperparameters in \equationref{eq:bernbeta} are then updated in closed form from the softened counts induced by $\{w_j\}$:
\begin{align}
\alpha_y &= \alpha_0 + \sum_{j} w_j\,\mathbb{I}(y_j{=}1), &
\beta_y  &= \beta_0 + \sum_{j} w_j\,\mathbb{I}(y_j{=}0), \nonumber
\end{align}
with the analogous update for the likelihood hyperparameters $(\alpha_{\pred|y}, \beta_{\pred|y})$ restricted to neighbors matching the conditioning value of $y$. We set the prior parameters by following two rules: 1) $\alpha_0 + \beta_0 = 1$; 2) they follow the prior statistical distributions from the $\mathcal{D}_{\mathrm{cal}}$.

\subsection{Details of Uncertainty Calibration}
\label{ap:calibration}

After quantifying uncertainty from neighborhood-based empirical correctness (see \secref{sec:quantification}), we further perform uncertainty calibration to ensure that the estimated probability accurately reflects the underlying true correctness probability.
Specifically, the goal of calibration is to refine the uncalibrated probability $\hat{p}_i$ into a calibrated probability $\tilde{p}_i$, such that
\begin{equation}
\Pr\big(y=1 \mid \tilde{p}_i = p\big) = p, \nonumber
\end{equation}
i.e., among all predictions assigned confidence $p$, the empirical frequency of correctness equals $p$.
This property ensures that the uncertainty estimate is statistically meaningful and can be reliably used for downstream decision making.

We perform calibration using the same held-out dataset $\mathcal{D}_{\mathrm{cal}}=\{(\hat{p}_i, \hat{y}_i, y_i)\}_{i=1}^{N}$, where $\hat{p}_i$ is the uncalibrated probability obtained from uncertainty quantification and $\mathbb{I}(y_i = \hat{y}_i)$ indicates whether the prediction is correct.
Calibration is formulated as learning a deterministic, monotonic mapping
\begin{equation}
g_\phi:[0,1]\rightarrow[0,1], \qquad \tilde{p}_i = g_\phi(\hat{p}_i), \nonumber
\end{equation}
such that the calibrated probabilities align with empirical correctness on $\mathcal{D}_{\mathrm{cal}}$.
In this work, we consider three widely used post-hoc calibration techniques~\citep{guo2017calibration}, each instantiating a different form of the mapping function $g_\phi$: 1) Temperature Scaling; 2) Isotonic Regression; and 3) Histogram Binning.

\paragraph{Histogram Binning.}
Histogram Binning is a non-parametric calibration method that partitions $[0,1]$ into $M$ disjoint bins $\{B_m\}_{m=1}^{M}$ and replaces each uncalibrated $\hat{p}_i$ by the empirical accuracy of its bin:
\begin{equation}
\tilde{p}_i
= g_{\mathrm{HB}}(\hat{p}_i)
= \frac{1}{|B_m|}
\sum_{j\in B_m} \mathbb{I}(y_j = \hat{y}_j),
\quad \text{if } \hat{p}_i \in B_m. \nonumber
\end{equation}
We include this method because it enforces calibration by construction with no parametric assumptions, providing a strong non-parametric baseline.
For this method, its piecewise-constant mapping can be sensitive to bin choices and data sparsity in low-density confidence regions.

\paragraph{Isotonic Regression.}
Isotonic Regression learns a non-parametric, monotonic calibration function by solving
\begin{align}
&\min_{g}
\sum_{i=1}^{N} \big(\mathbb{I}(y_i = \hat{y}_i) - g(\hat{p}_i)\big)^2\quad\text{s.t. } g(p) \ge g(q) \quad \forall\, p \ge q. \nonumber
\end{align}
It preserves the ranking of $\hat{p}_i$ monotonically yet flexibly adjusts bin sizes to minimize squared calibration error, producing a smoother mapping than Histogram Binning.

\paragraph{Temperature Scaling.}
Temperature Scaling is a parametric calibration method that rescales the logit corresponding to each uncalibrated probability by a single scalar $\tau>0$:
\begin{equation}
\tilde{p}_i = g_{\mathrm{TS}}(\hat{p}_i) = \sigma\!\left(\frac{1}{\tau}\log\frac{\hat{p}_i}{1-\hat{p}_i}\right), \nonumber
\end{equation}
where $\sigma(\cdot)$ is the sigmoid.
The temperature $\tau$ is optimized on $\mathcal{D}_{\mathrm{cal}}$ by minimizing negative log-likelihood.

\subsection{Details of Interactive Policy Formulations}
\label{ap:policy}

We show details of the conformal-prediction policy and model-based policy in this section, which make decisions based on the tracked uncertainties to generate actions.

\subsubsection{Model-based Policy}
\label{ap:planning}

The model-based policy formulates object-oriented manipulation under uncertainty as a sequential decision-making problem, where the robot maintains a belief over states and plans actions to maximize cumulative reward while trading off information gathering against physical execution.

\textbf{State and Belief}:
Each state $s = s^o \cup s^r$ is object-oriented, where $s^o = (s_1, \ldots, s_n)$ collects factorized per-object states and $s^r = \{s_{ij}\}$ encodes pairwise blocking relations.
Each object state $s_i$ is a vector of boolean predicates $\{\lambda_j(i)\}$ that expand across multi-turn interactions.
We adopt this object-level factorization because the joint state space grows as $2^n$ in the number of objects, whereas factorized inference scales linearly and matches the object-centric structure of grounding predictions.
The robot maintains a factorized belief $b(s)=\prod_i b(s_i)$ updated by Bayes' rule from observations.

\textbf{Action Space}:
The action space consists of four types: $\acti(q)$ asks an open-ended question to globally reduce uncertainty; $\actc(\obj_i)$ confirms whether object $\obj_i$ is the target; $\actg(\obj_i)$ grasps object $\obj_i$ as the target; $\actr(\obj_i)$ removes object $\obj_i$ as a blocking obstacle.
For $\acti$ we prompt a language model with the current scene to propose a single informative question (Prompt~\ref{prompt:informative_question}), reducing the otherwise infinite question space to $|\mathcal{Q}|=1$. We use an LM-proposed single candidate because exhaustive search over open-ended natural-language questions is intractable, and a context-conditioned proposal already captures most of the value of asking.

\textbf{Observation Model}:
Instead of modeling raw observation distributions over images and language, we use the calibrated visual grounding predictions directly as observations, since they are already calibrated by \secref{sec:uncertainty} to satisfy $\Pr(y\!=\!1\mid\tilde{p})=\tilde{p}$ and serve as a sufficient summary of the raw image for the binary target/non-target decision.
For each object $i$, the observation model is $Z(\hat{y}_i \mid s_i = y_i, a) = \tilde{p}_i$.
For linguistic observations from $\acti$, the model follows a voting formulation: $Z(u \mid s, a) \propto \sum_{i=1}^{n} \mathbb{I}(s_i \text{ matches } u)$, where $u$ is the user's answer.

\textbf{Transition Model}:
For interactive actions ($\acti$, $\actc$), the physical state is unchanged: $T(s' \mid s, a) = \mathbb{I}(s' = s)$.
$\acti$ actions that introduce new predicates will expand the state space accordingly.
For physical actions ($\actr$, $\actg$), the transition removes the manipulated object and its blocking edges from the object graph.
This deterministic model is appropriate because manipulation success is observable post-hoc and updates belief rather than as transition stochasticity.
Therefore, it is a simple approximation that makes planning tractable.

\textbf{Reward}:
The reward function balances task completion, obstacle removal, and interaction cost, computed in expectation over the current belief.
Grasping a target yields $+5$ and mistakenly grasping a non-target or removing a target incurs $-5$; removing an object that blocks $k$ potential targets receives $+3k$; removing a non-blocking, non-target object incurs $-1$; interactive actions ($\acti$, $\actc$) each incur $-2$.

\textbf{Planning Objective}:
The policy aims to find an action sequence that maximizes expected cumulative reward over a finite horizon $T$, matching the main text:
$\pi^* = \arg\max_\pi \mathbb{E}\!\left[ \sum_{t=0}^{T} r(s_t, a_t) \right]$.
We solve this online with a depth-limited forward search (depth up to 3 steps) that expands the factorized belief, propagating it through the per-object observation model.

\subsubsection{Conformal-Prediction Policy}
\label{ap:conformal}

The conformal-prediction policy follows \cite{ren2023robots} and turns the calibrated probabilities $p_{\obj,\step}$ from \secref{sec:uncertainty} into prediction sets with a statistical coverage guarantee.

The calibration procedure on $\mathcal{D}_{\mathrm{cal}} = \{(I_j, \expr_j, r_j, y_j)\}_{j=1}^{N}$ proceeds as follows:
\begin{itemize}
    \item For each calibration sample, extract the score assigned to the ground-truth correct object.
    \item Sort these scores in descending order to obtain $\{s_{(1)}, s_{(2)}, \ldots, s_{(N)}\}$.
    \item For each candidate threshold $\tau_k = s_{(k)}$, compute the empirical miscoverage $\hat{e}_k$ and $v_k = \lfloor (N+1) \hat{e}_k \rfloor$.
    \item Compute the coverage lower bound from a Beta tail: $\beta_k = F^{-1}_{\mathrm{Beta}(N+1-v_k,\, v_k)}(\delta)$, where $\delta\in(0,1)$ is the calibration confidence level controlling how likely the guarantee may fail given the finite calibration set $\mathcal{D}_{\mathrm{cal}}$. We use $\delta = 0.1$, so the coverage guarantee holds with probability at least $0.9$.
    \item For target coverage $1-\alpha$, select the smallest $k$ such that $\beta_k \geq 1-\alpha$, and set $\tau(\alpha) = s_{(k)}$.
\end{itemize}
By inverting a Beta tail rather than using the empirical quantile, this calibration statistically accounts for finite-sample uncertainty in $\mathcal{D}_{\mathrm{cal}}$.

At test time the calibrated threshold $\tau(\alpha)$ is used inside two action policies: the \emph{image-wise} and \emph{object-wise} conformal policies.
Both of them are described concretely in \apref{ap:ivg_baseline_details}.
They share the same calibrated threshold but differ in whether the conformal set ranges over candidate objects in a single image or over the binary $\{$True, False$\}$ label of a single object.

\subsubsection{Details of Hierarchical Planning in Clutter}
\label{ap:hierarchy}
While the interaction policy selects actions to reduce uncertainty, all actions must also be physically executable by the robot.
We adopt a hierarchical planning framework that jointly considers uncertainty and motion constraints, following the idea of Task and Motion Planning (TAMP) \citep{kaelbling2011hierarchical, garrett2021integrated}.
We reason symbolically and validate only the next physical action because full TAMP search is intractable at interaction time, and a per-step feasibility check is sufficient when the symbolic blocking graph is updated reactively.
At step $\step$, we define a feasible action set
$\hat{\mathcal{A}}_\step=\{a\in\act\mid \mathcal{F}(a,\step)=1\}$,
where $\mathcal{F}(a,\step)$ indicates whether action $a$ is executable.
For $a\in\acti$, $\mathcal{F}(a,\step)\equiv 1$.
For object-specific actions $a\in\actc\cup\actr\cup\actg$, feasibility requires that the target object is not physically blocked.
In particular, although $\actc$ is a linguistic action, confirming a blocked object is not allowed because the robot cannot visually disambiguate the occluded candidate, so the answer would be unreliable.
If an object is blocked, the planner must first remove the blocking objects.
Feasibility further requires collision-free motion planning.
To approximate blocking relations efficiently, we initialize the blocking matrix $\{s_{ij}\}_{i,j=1}^{n}$ from a visual relation model \citep{zhang2018visual}.
Actions violating $s_{ij}$ are pruned within policies.
If a physical action fails (e.g., motion-planning failure or unreachable grasp), newly discovered blockers are added to $s_{ij}$ and the system replans.
The resulting bi-level planner integrates symbolic reasoning with physical feasibility as an efficient approximation to integrated task-motion planning in cluttered environments.

\subsubsection{Derivation of Uncertainty Tracking}
\label{ap:derv_uncert_track}
We track object-centric uncertainty over steps and update it sequentially after each interaction.
At time step $\step$, given the current visual observation $I_t$ and the language history $\expr_{0:\step}$, our goal is to estimate
\[
p_{\obj,\step}
=
p(y=1 \mid I_t,\expr_{0:\step},\obj),
\]
which denotes the probability that region $\obj$ is the true target conditioned on $I_t$ and all language observations.
We now derive the recursive decomposition of $p_{r,t}$.
Start from the definition and apply Bayes' rule:
\begin{align}
p_{r,t}&=p(y=1 \mid I_t,\expr_{0:\step},\obj)\propto
p(\expr_\step \mid y=1, I_t,\expr_{0:\step-1},\obj)\;
p(y=1 \mid I_t,\expr_{0:\step-1},\obj).
\label{eq:update_step1}
\end{align}
Assume conditional independence of the new expression given $(y,I_t,\obj)$:
\begin{equation}
p(\expr_\step \mid y, I_t,\expr_{0:\step-1},\obj)
=
p(\expr_\step \mid y, I_t,\obj).
\label{eq:cond_indep}
\end{equation}
Substituting \eqref{eq:cond_indep} into \eqref{eq:update_step1} gives
\begin{equation}
p(y\mid I_t,\expr_{0:\step},\obj)
\propto
p(\expr_\step \mid y, I_t,\obj)\;
p(y\mid I_t,\expr_{0:\step-1},\obj).
\label{eq:update_step2}
\end{equation}
Using Bayes' rule again:
\begin{equation}
p(y\mid I_t,\expr_\step,\obj)
=
\frac{
p(\expr_\step \mid y, I_t,\obj)\;p(y\mid I_t,\obj)
}{
p(\expr_\step \mid I_t,\obj)
}, \nonumber
\end{equation}
which implies
\begin{equation}
p(\expr_\step \mid y, I_t,\obj)
\propto
\frac{
p(y\mid I_t,\expr_\step,\obj)
}{
p(y\mid I_t,\obj)
}. \nonumber
\end{equation}
Substituting into \eqref{eq:update_step2}, we obtain the recursive update
\begin{align}
p_{\obj,\step}
\propto \eta\frac{\;p(y=1\mid I_t,\expr_\step,\obj)}{p(y=1\mid I_t,\obj)}\;p_{\obj,\step-1},  \nonumber
\end{align}
where $\eta$ is the normalizer.

\subsection{Additional Details of Visual Grounding Benchmarking}
\label{ap:bench_result}

\subsubsection{Evaluation Metrics}
\label{ap:cali_metrics}

We evaluate uncertainty estimation using three categories of metrics: task performance, calibration quality, and discrimination ability.
For each prediction $i$, we denote the predicted label as $\hat{y}_i \in \{0,1\}$, the ground-truth label as $y_i \in \{0,1\}$, and the calibrated confidence as $\tilde{p}_i \in [0,1]$.

\paragraph{Task Performance}
\textit{Recall@K} measures the fraction of samples for which the ground-truth target appears among the top-$K$ predictions ranked by confidence.
In particular, \textit{Recall@1} corresponds to standard grounding accuracy:
\begin{equation}
\mathrm{Recall@1} = \frac{1}{N}\sum_{i=1}^{N} \mathbb{I}(\hat{y}_i = y_i), \nonumber
\end{equation}
where a prediction is counted as correct if its predicted bounding box has an Intersection-over-Union (IoU) greater than $0.5$ with the ground-truth box.

\paragraph{Calibration Metrics}
Calibration assesses whether predicted confidences align with empirical correctness frequencies.
We partition predictions into $M$ bins according to their confidence levels.
For the $m$-th bin, let $B_m$ denote the set of samples whose confidence falls in that bin, and define
\[
\mathrm{acc}(B_m) = \frac{1}{|B_m|}\sum_{i \in B_m} \mathbb{I}(\hat{y}_i = y_i), \quad
\mathrm{conf}(B_m) = \frac{1}{|B_m|}\sum_{i \in B_m} \tilde{p}_i.
\]
\begin{itemize}
    \item \textit{Expected Calibration Error (ECE)} quantifies the weighted average gap between confidence and accuracy:
    \[
    \mathrm{ECE} = \sum_{m=1}^{M} \frac{|B_m|}{N} \, \big|\mathrm{acc}(B_m) - \mathrm{conf}(B_m)\big|.
    \]
    ECE uses fixed-width bins (e.g., $[0, 0.1), [0.1, 0.2), \ldots$).
    \item \textit{Adaptive ECE (aECE)} follows the same formulation but uses variable-width bins, where each bin contains an equal number of samples. This reduces sensitivity to bin boundaries.
    \item \textit{Negative Log-Likelihood (NLL)} measures the log-probability assigned to the correct outcome:
    \[
    \mathrm{NLL} = -\frac{1}{N}\sum_{i=1}^{N} \big[ y_i \log \tilde{p}_i + (1 - y_i) \log (1 - \tilde{p}_i) \big].
    \]
    Lower NLL indicates better probabilistic predictions.
    \item \textit{Brier Score} is the mean squared error between predicted confidence and true correctness:
    \[
    \mathrm{Brier} = \frac{1}{N}\sum_{i=1}^{N} (\tilde{p}_i - y_i)^2.
    \]
    It is a proper scoring rule that jointly captures calibration and sharpness.
    \item \textit{Reliability diagrams} visualize calibration by plotting $\mathrm{acc}(B_m)$ versus $\mathrm{conf}(B_m)$ for each bin $m$. Perfect calibration corresponds to the diagonal $y = x$.
\end{itemize}

\paragraph{Discrimination Metrics}
Discrimination assesses the ability to distinguish correct predictions from incorrect ones using uncertainty scores.
We adopt two standard metrics from binary classification:
\begin{itemize}
    \item The \textit{Receiver Operating Characteristic (ROC)} curve plots the true positive rate against the false positive rate as the decision threshold varies.
    The \textit{Area Under the ROC Curve (AUROC)} summarizes this trade-off as a single scalar, representing the probability that a randomly selected error receives higher uncertainty than a randomly selected correct prediction.
    An AUROC of $1.0$ indicates perfect separation, while $0.5$ corresponds to random ranking.
    \item The \textit{Precision–Recall Curve (PRC)} plots precision against recall as the threshold varies.
    The \textit{Area Under the PRC (AUPRC)} is particularly informative under class imbalance (e.g., when most predictions are correct), as it focuses on the minority class (errors) without being dominated by the majority class.
\end{itemize}

\begin{figure}[htbp]
    \centering
    \subfloat[Ablation of Retrieval Strategy]{
        \includegraphics[width=1.0\textwidth]{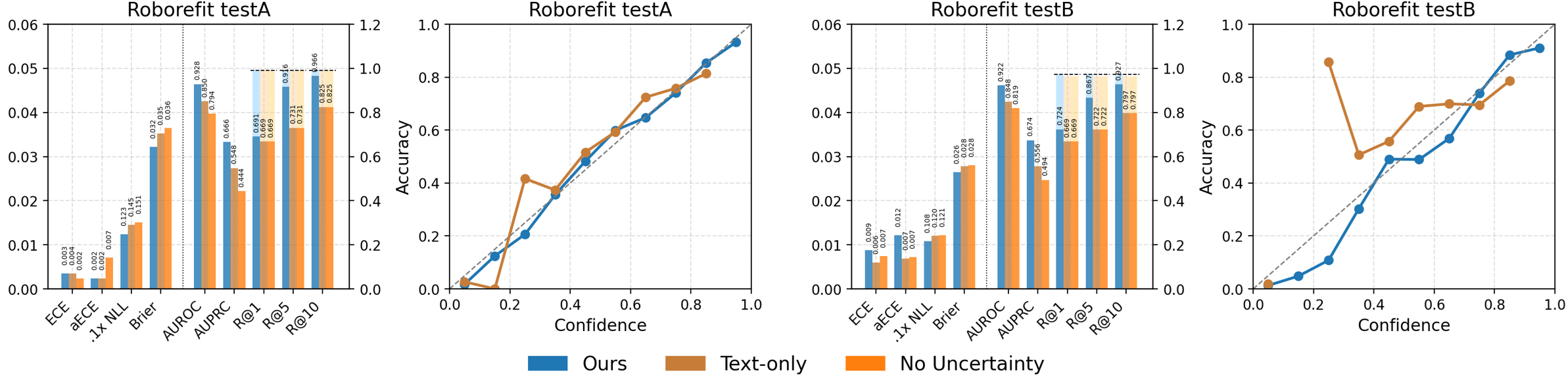}
        \label{fig:retrieval_ablation_detail}
    } \\
    \subfloat[Ablation of Calibration Methods]{
        \includegraphics[width=1.0\textwidth]{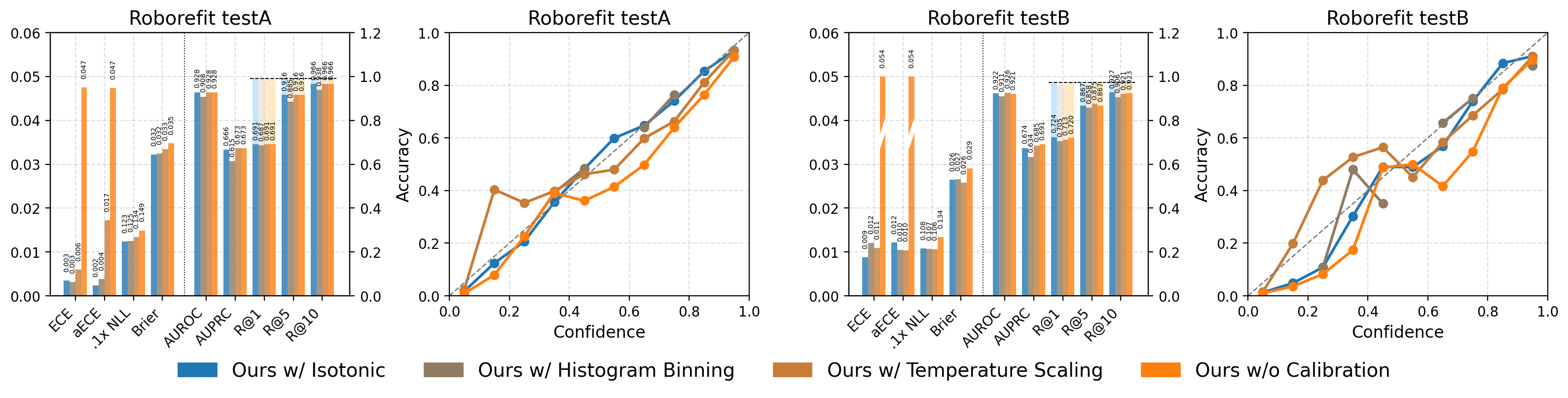}
        \label{fig:calib_ablation_detail}
    } \\
    \subfloat[Ablation of Held-out Data Scale]{
        \includegraphics[width=1.0\textwidth]{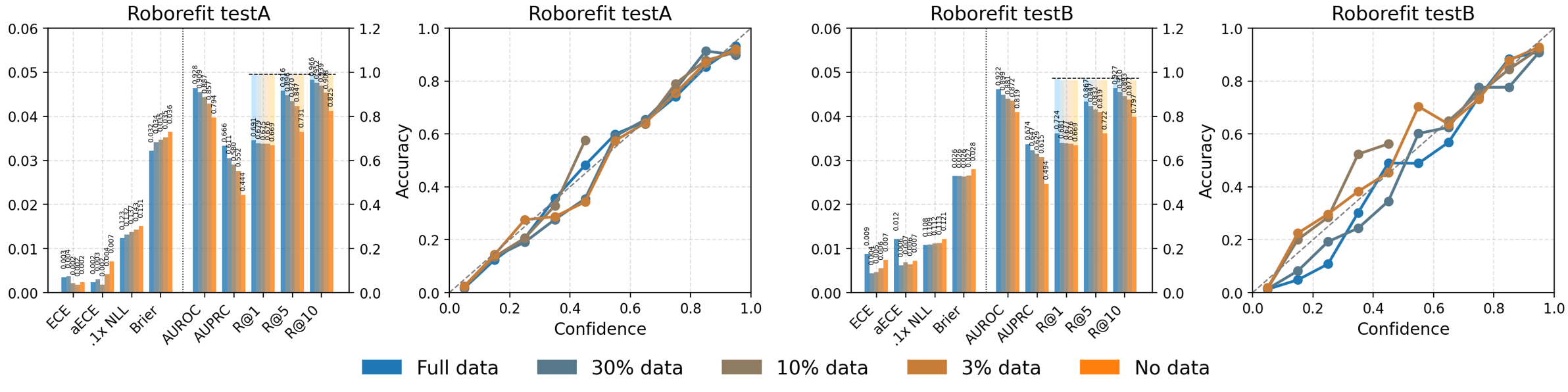}
        \label{fig:data_scale_ablation_detail}
    }
    \caption{
    Ablation studies on uncertainty estimation. We analyze the effect of (a) different retrieval strategies for uncertainty quantification, (b) different post-hoc calibration methods, and (c) varying held-out data scales on uncertainty quality.
    }
    \label{fig:ablation_studies}
\end{figure}

\subsubsection{Ablation on Retrieval Strategy}
\label{ap:retrieval_ablation}
For the uncertainty quantification part, we show detailed comparison of our full three-stage retrieval against two representative baselines that ablate the retrieval design while keeping everything else identical:
\begin{itemize}
    \item \textbf{No-Uncertainty (No-U):} uses the raw model outputs as the confidence without uncertainty, isolating the effect of adding any surrogate at all.
    \item \textbf{Language-based Uncertainty (Lang-U):} estimates uncertainty with $\mathcal{N}(h)$ retrieved using only text queries. This isolates the contribution of cross-modal and visual stages over a text-only neighborhood.
\end{itemize}
We evaluate Calibration (ECE, aECE, NLL, Brier), Discrimination (AUROC, AUPRC), and Task Performance (Recall@K).
Results in \figref{fig:retrieval_ablation_detail} demonstrate that our method outperforms on all three metric families.
Concretely, higher discrimination metrics show that our method distinguishes between correct and incorrect predictions with a clearer margin, benefiting from the multi-modal neighborhoods.
Notably, the Recall@K results show that text-only retrieval provides no improvement for object-centric predictions.
With a single text query, it cannot distinguish correct predictions from incorrect ones within the same image.
This happens because the black-box model produces deterministic outputs, so using only the text input gives the same uncertainty scores for all objects.
Another issue of text-only retrieval is that it imposes a loose standard for the retrieved samples, and hence tends to produce uncertainty scores on average, which can be observed from the reliability diagrams in \figref{fig:retrieval_ablation_detail}.
The estimated uncertainty tends to concentrate on two separated regions, which reveals that the uncertainty is degrading to the average on the dataset.
By contrast, our method achieves smooth prediction and distinguishes different samples well, which aligns better to the empirical correctness.

\subsubsection{Ablation on Uncertainty Calibration}
\label{ap:calibration_ablation}

We further evaluate whether post-hoc calibration is necessary to obtain numerically accurate uncertainty estimates after quantification.
Although we confirm that our quantification module produces well-ranked uncertainty scores, they may still be biased by the error from neighborhood retrieval.
To validate this, we first compute the raw uncertainties from our uncertainty quantification method and then apply a range of calibration techniques, including Histogram Binning (\textit{HB}), Isotonic Regression (\textit{IR}), and Temperature Scaling (\textit{TS}).
We visualize reliability diagrams and report metrics of both \textit{calibration} and \textit{discrimination}.
This analysis examines how uncertainty calibration affects the numerical accuracy of uncertainty as well as quality of ranking positive samples against negative samples.
The results in \figref{fig:calib_ablation_detail} show that post-hoc calibration is necessary to get numerically accurate uncertainty estimates, even when the underlying uncertainty scores are well ranked.
High calibration errors and the oscillating reliability diagrams of \textit{Ours w/o Calibration} on both testing sets confirm that the bias induced by neighborhood retrieval can significantly affect the fidelity of the numerical uncertainty values.
By contrast, all post-hoc calibration methods can consistently improve reliability across both test splits, aligning predicted confidence with empirical correctness.
Comparing different post-hoc calibration methods, \textit{Ours w/ Isotonic} achieves the most consistent calibration performance, yielding the lowest calibration errors and the smoothest reliability curves.
It also preserves uncertainty ranking quality well.
\textit{Ours w/ Histogram Binning} and \textit{Ours w/ Temperature Scaling} also reduce miscalibration but are less stable, particularly in low-density, mid-confidence regions.
Results indicate that our uncertainty quantification provides reasonably well-ranked uncertainty that is not numerically accurate enough, while post-hoc calibration is essential to make these uncertainties interpretable and trustworthy.

\subsubsection{Additional Results of Input Perturbations}
\label{ap:input_ablation}

To systematically analyze how visual and linguistic noise affect uncertainty estimation, we construct controlled noise conditions that reflect common sources of variation in realistic robotic applications.
We evaluate on a 10\% random subset of the full test sets because the perturbation grid covers many noise types and intensities, and running each VLM at every combination on the full set is prohibitive.
We consider four perturbation types: \textit{Gaussian} and \textit{Occlusion} as visual noise, \textit{Typo} and \textit{Rephrase} as linguistic perturbations.
\textit{Gaussian} simulates photometric noise from hardware or poor lighting and \textit{Occlusion} masks parts of the image to represent clutter; \textit{Typo} introduces character-level noise and \textit{Rephrase} uses a large language model (Qwen2.5 72B) to generate uncommon but semantically equivalent expressions. The four were chosen to span the two input modalities and to mix low-level signal corruption with high-level semantic shift, so that the same uncertainty estimator is stressed along disjoint noise axes.

\begin{figure}[t]
    \centering
    \begin{minipage}[t]{0.48\textwidth}
        \centering
        \includegraphics[width=\textwidth]{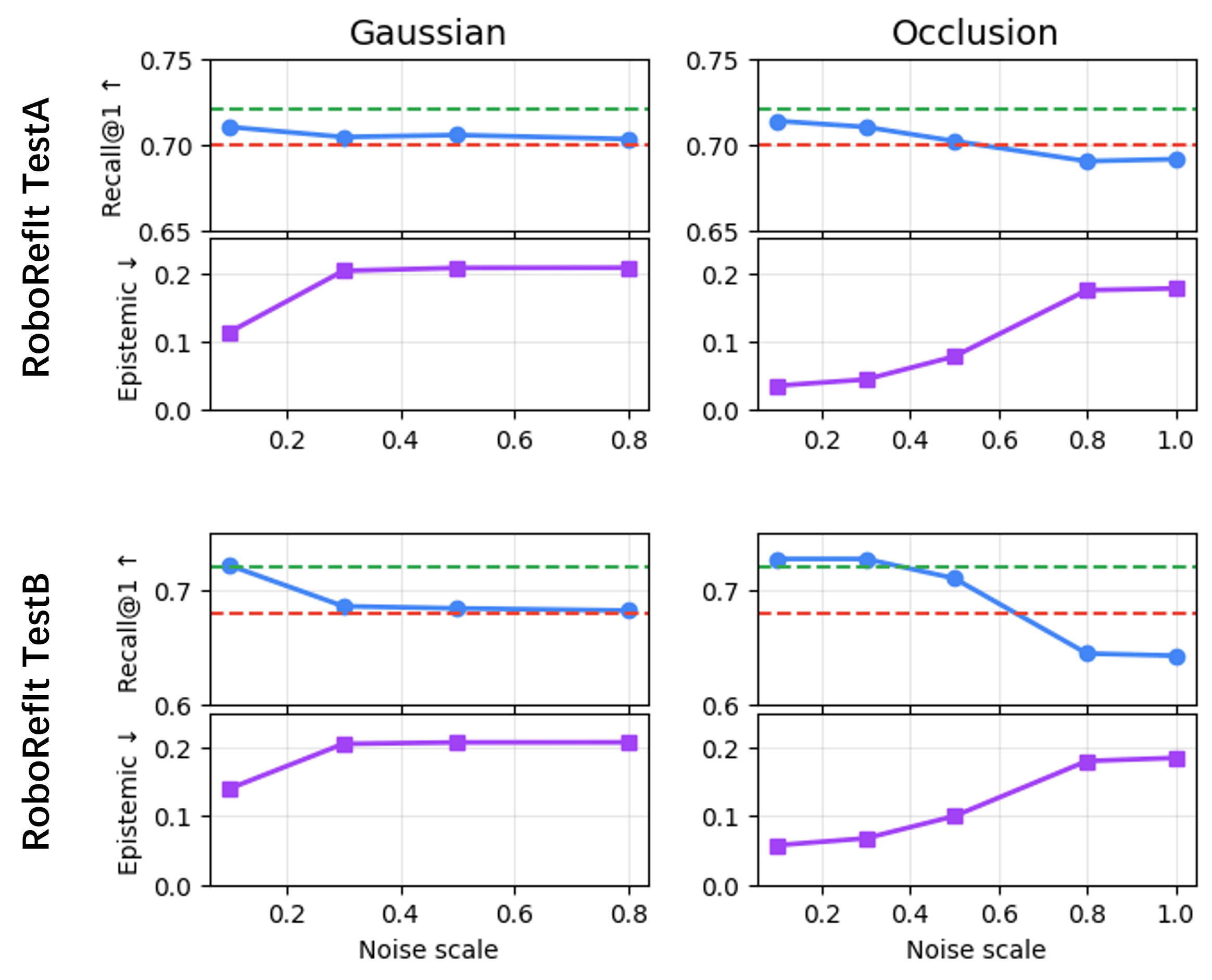}
        \caption{
            Model performance gain from uncertainty estimation and epistemic uncertainty versus visual noise. Gaussian and occlusion refer to adding different scales of Gaussian noise and black padding to emulate different observation conditions.
        }
        \label{fig:visual_noise}
    \end{minipage}
    \hfill
    \begin{minipage}[t]{0.48\textwidth}
        \centering
        \includegraphics[width=\textwidth]{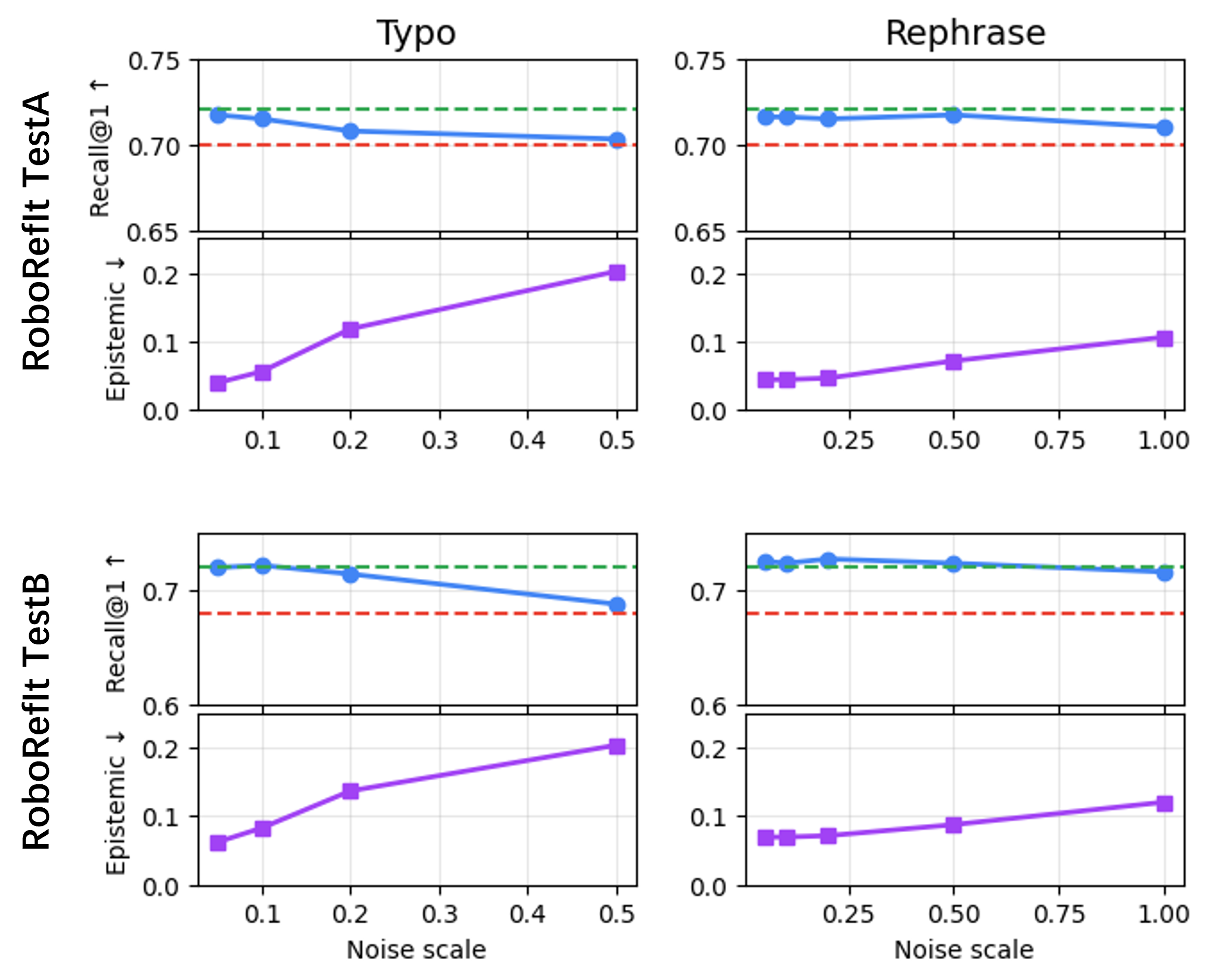}
        \caption{
            Model performance gain from uncertainty estimation and epistemic uncertainty versus textual noise. ``Typo'' means randomly replacing chars with random chars, doubling itself, or removing it. ``Rephrase'' means calling a large language model to rephrase the input text with uncommon words.
        }
        \label{fig:text_noise}
    \end{minipage}
\end{figure}

To isolate the effect of noise on uncertainty estimation, we keep the upstream object detection and grounding predictions fixed (generated from original inputs) and only inject noise into the uncertainty estimation stage.
\figref{fig:visual_noise} and \figref{fig:text_noise} show the results.
We observe that increasing noise leads to a gradual degradation in task performance, accompanied by a corresponding increase in estimated uncertainty.
Crucially, there is a consistent rise in \textit{epistemic} uncertainty across all noise types, confirming that it effectively captures the increasing out-of-distribution (OOD) shift caused by the perturbations.

\subsubsection{Qualitative Examples of Uncertainty Decomposition}
\label{ap:uncert_qual}
\begin{figure}[t]
    \centering
    \includegraphics[width=\textwidth]{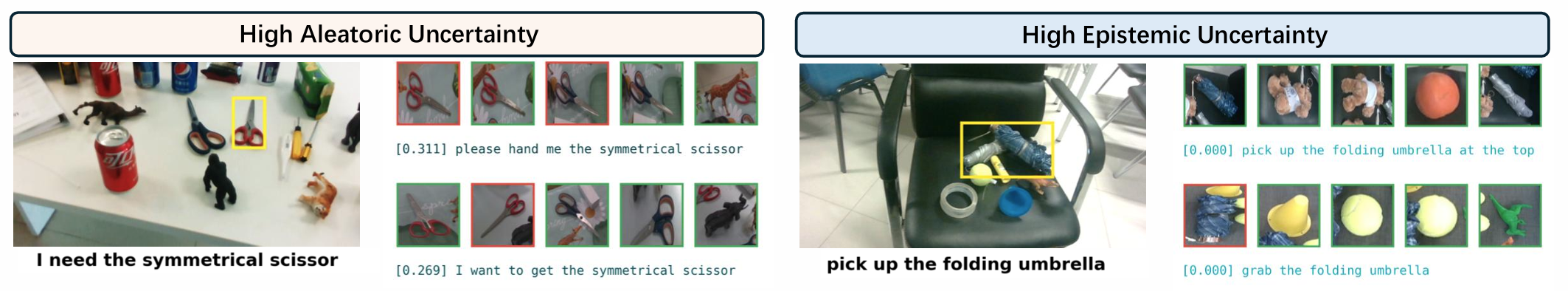}
    \caption{Qualitative examples of high \textit{aleatoric} and high \textit{epistemic} uncertainty. Top-retrieved neighbors are shown with mean similarity on the left. Red boxes mark model-predictive errors.}
    \label{fig:uncert_qual}
\end{figure}

\figref{fig:uncert_qual} visualizes the aleatoric/epistemic decomposition on representative cases.
High \textit{aleatoric} uncertainty arises when neighborhoods are dense in $\mathcal{D}_{\mathrm{cal}}$ but predictions frequently disagree with labels (systematic VLM failure rather than per-sample noise).
High \textit{epistemic} uncertainty arises when neighborhoods are sparse and semantically divergent, i.e., a genuinely out-of-distribution query.

\subsubsection{Ablation on Cross-Dataset Transfer}
\label{ap:cross_dataset}

\begin{table}[t]
\centering
\caption{Cross-dataset validation (Recall@1~/~AUPRC) on Molmo. Rows choose the held-out calibration source for our surrogate. Columns evaluate on each test distribution. The row \textbf{None} reports the raw VLM without our uncertainty estimation (i.e., no surrogate at all), serving as the no-uncertainty floor. \textbf{\underline{Best}} and \textbf{Second Best} per column.}
\label{tab:cross_dataset}
\resizebox{\columnwidth}{!}{%
\begin{tabularx}{1.15\columnwidth}{l *{6}{>{\centering\arraybackslash}X}}
\toprule
\multirow{2}{*}{\textbf{Held-Out}}
& \multicolumn{2}{c}{\textbf{RefCOCO}}
& \multirow{2}{*}{\textbf{RefCOCOg}}
& \multicolumn{2}{c}{\textbf{RoboRefIt}}
& \multirow{2}{*}{\textbf{Average}} \\
\cmidrule(lr){2-3}\cmidrule(lr){5-6}
& {\textbf{Test A}} & {\textbf{Test B}} & & {\textbf{Test A}} & {\textbf{Test B}} \\
\midrule
{\textbf{None}}
& 78.3/69.5 & 67.7/41.5 & 66.1/46.8 & 66.9/44.4 & 66.9/49.4 & 69.2/50.3 \\
{\textbf{RefCOCO}}
& {\textbf{\underline{79.6}}}/{\textbf{\underline{80.9}}} & {\textbf{\underline{70.2}}}/{\textbf{57.2}} & {\textbf{\underline{69.7}}}/{\textbf{59.5}} & 68.1/53.2 & 68.7/55.7 & {\textbf{71.2}}/{\textbf{\underline{61.3}}} \\
{\textbf{RefCOCOg}}
& {\textbf{79.5}}/{\textbf{77.3}} & {\textbf{69.6}}/{\textbf{\underline{57.9}}} & {\textbf{69.5}}/{\textbf{\underline{59.6}}} & 67.6/52.0 & 67.7/54.2 & 70.8/60.2 \\
{\textbf{RoboRefIt}}
& 77.5/69.3 & 68.8/44.9 & 65.5/46.8 & {\textbf{\underline{69.1}}}/{\textbf{\underline{66.6}}} & {\textbf{\underline{71.3}}}/{\textbf{\underline{67.1}}} & 70.4/58.9 \\
{\textbf{VG3000}}
& 79.2/74.8 & 69.5/51.1 & 69.0/55.9 & {\textbf{68.2}}/{\textbf{58.8}} & {\textbf{70.5}}/{\textbf{63.1}} & {\textbf{\underline{71.3}}}/{\textbf{60.7}} \\
\bottomrule
\end{tabularx}%
}
\end{table}

\tabref{tab:cross_dataset} shows the cross-domain validation performance.
We observe a consistent dependency of uncertainty quality on the choice of the held-out calibration dataset.
When the held-out dataset matches the testing distribution, the model usually achieves the best performance.
Higher epistemic uncertainty systematically corresponds to lower performance across testing datasets, indicating that our uncertainty estimator reliably captures distributional mismatch.
Notably, even in the worst case, a nearly out-of-distribution held-out dataset does not obviously harm performance, serving as a stable lower bound.
The \textit{VG3000} subset often achieves competitive second-best performance with moderate uncertainty, suggesting that a semantically diverse dataset can serve as a robust alternative when in-domain data are unavailable.

\subsubsection{Ablation on Data Scale}
We also vary the percentage of held-out data (3\%, 10\%, 30\%, 100\%) to understand how data scale impacts uncertainty quality, as shown in \figref{fig:data_scale_ablation_detail}.
Results demonstrate that uncertainty estimation remains effective even with very limited held-out data (e.g., 3\%, approximately 1000 images).
With larger held-out datasets, discrimination performance improves monotonically, while calibration improvements are more incremental.
This highlights the scalability and practicality of our method: a small, carefully collected held-out dataset is sufficient to produce reliable uncertainty estimation for real-world deployment.

\begin{figure}[t]
    \centering
    \subfloat[Discrimination Performance Details]{
        \includegraphics[width=1.0\textwidth]{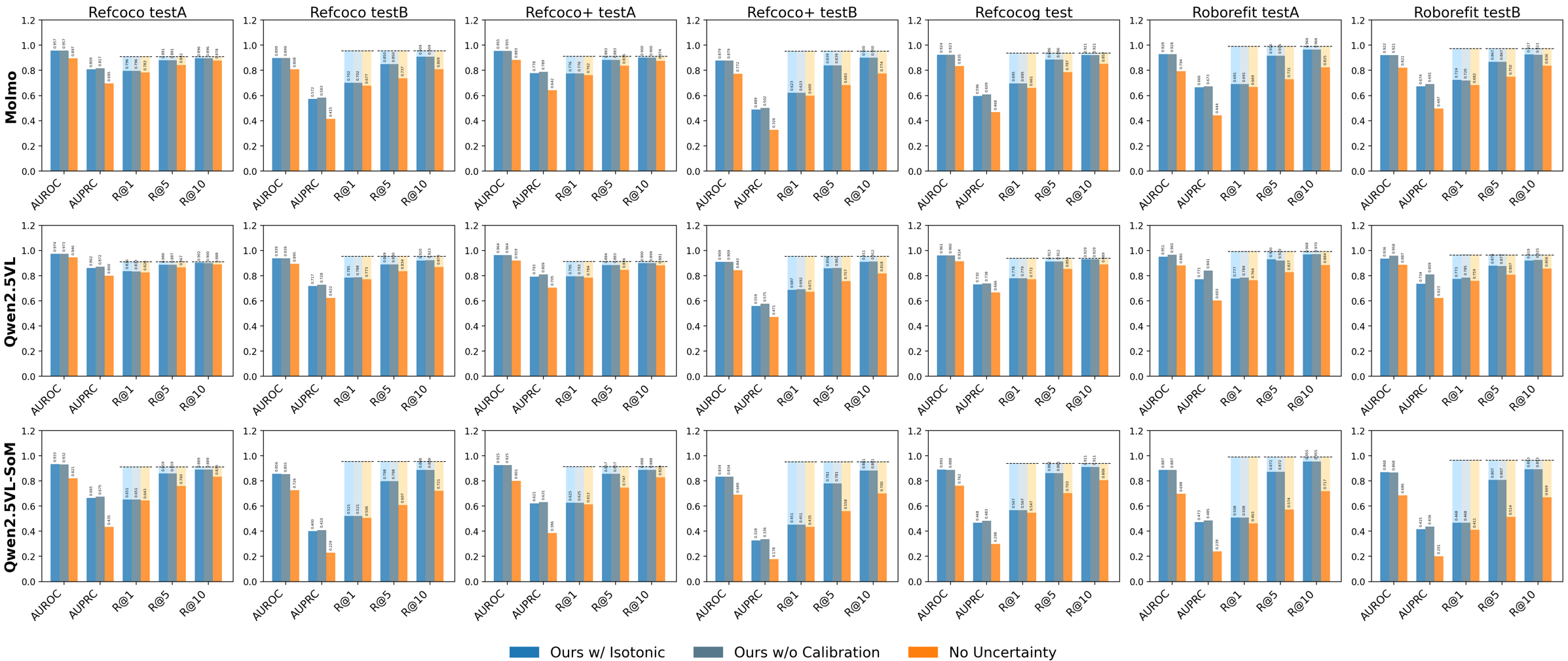}
        \label{fig:disc_detail}
    } \\
    \subfloat[Calibration Performance Details]{
        \includegraphics[width=1.0\textwidth]{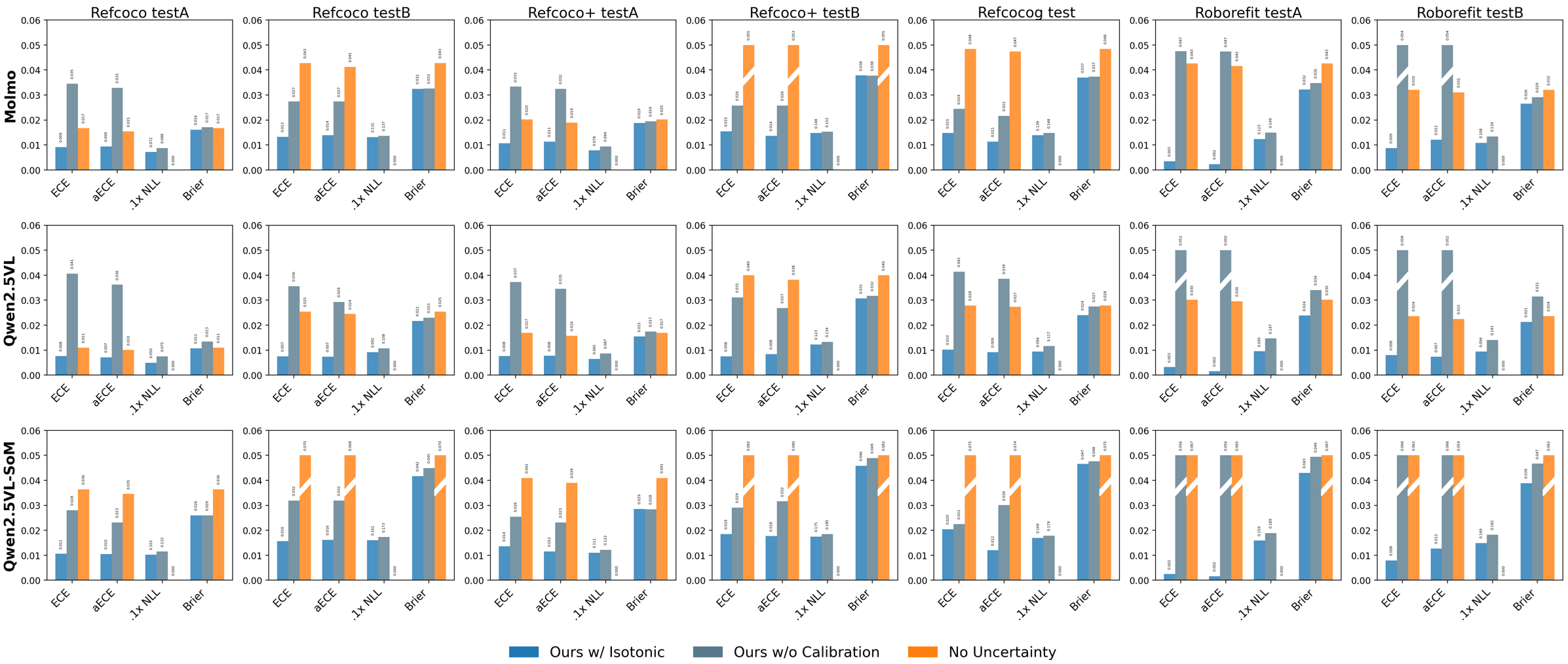}
        \label{fig:calib_detail}
    }
    \caption{
    Detailed uncertainty performance across datasets and visual grounding models. We report (a) discrimination metrics and (b) calibration metrics for each model-dataset combination.
    }
    \label{fig:detailed_performance}
\end{figure}

\begin{figure}[htbp]
    \centering
    \subfloat[Reliability Diagram without Uncertainty Calibration]{
        \includegraphics[width=0.95\textwidth]{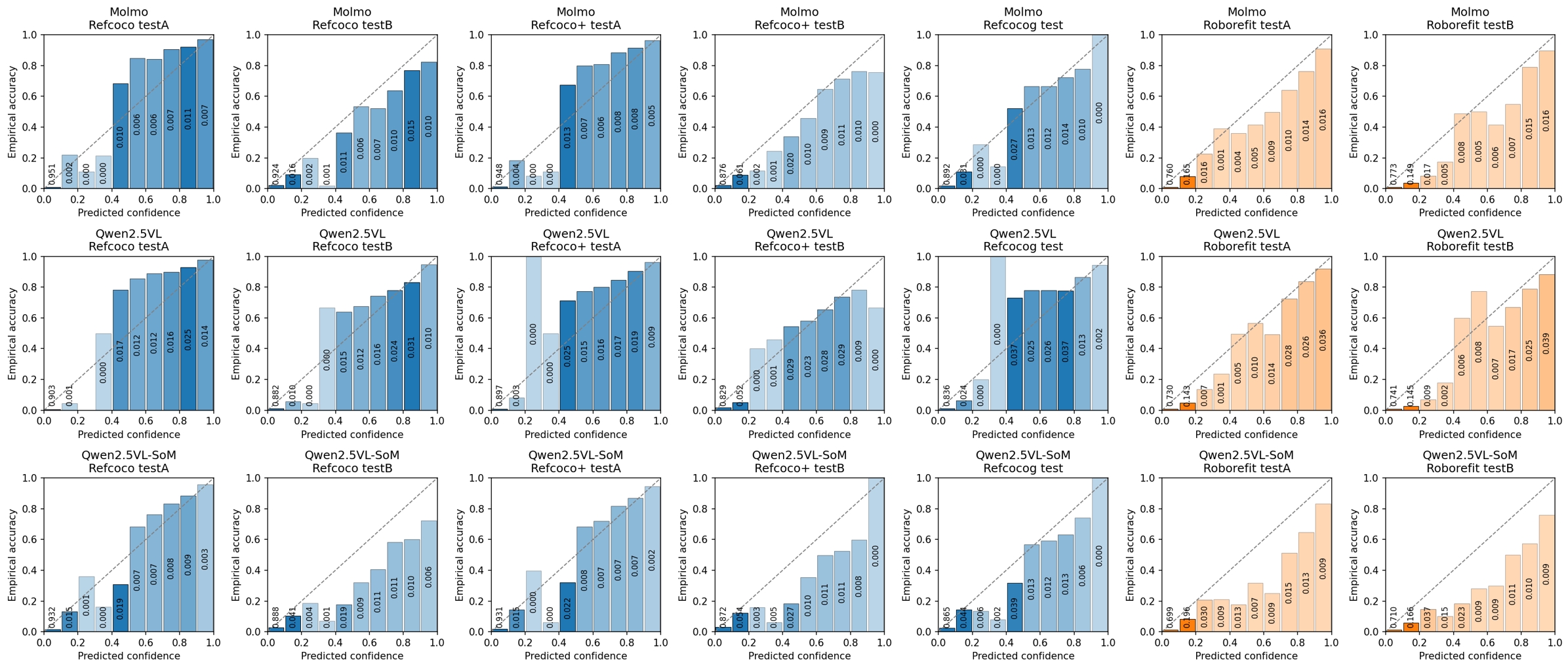}
        \label{fig:no_cal}
    } \\
    \subfloat[Calibrators from train/validation dataset]{
        \includegraphics[width=0.95\textwidth]{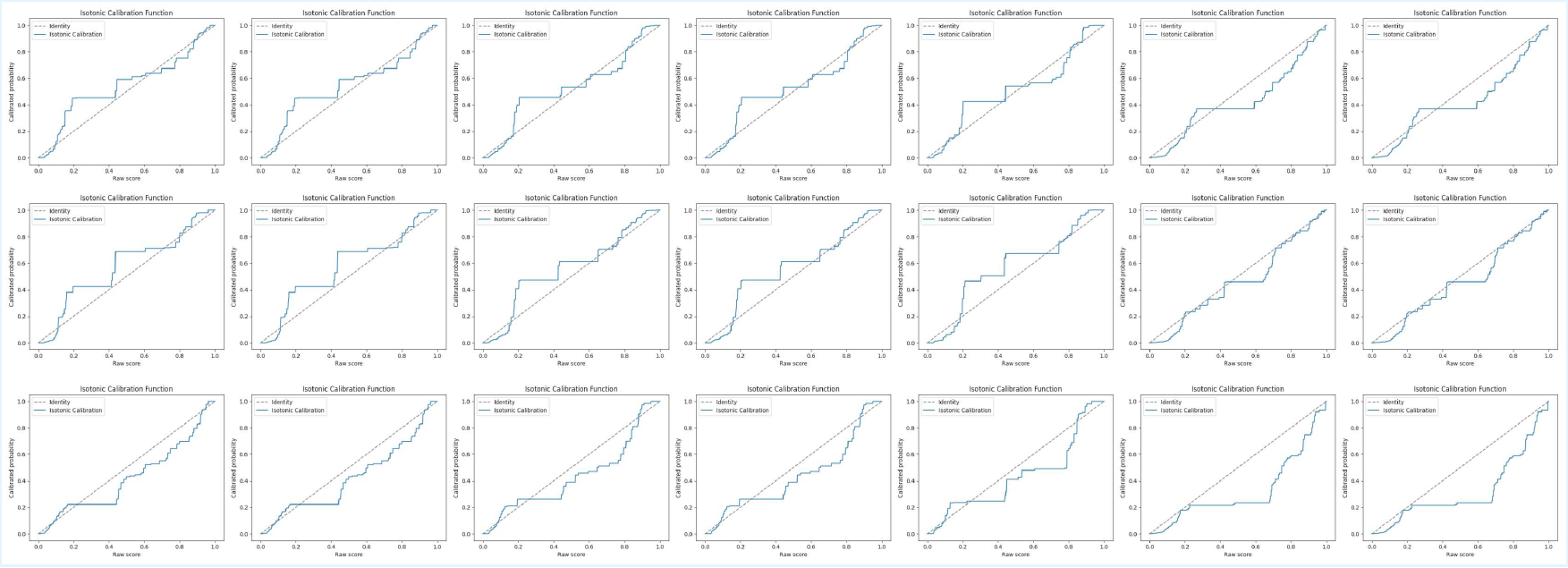}
        \label{fig:calibrator}
    } \\
    \subfloat[Reliability Diagram with Post-hoc Uncertainty Calibration]{
        \includegraphics[width=0.95\textwidth]{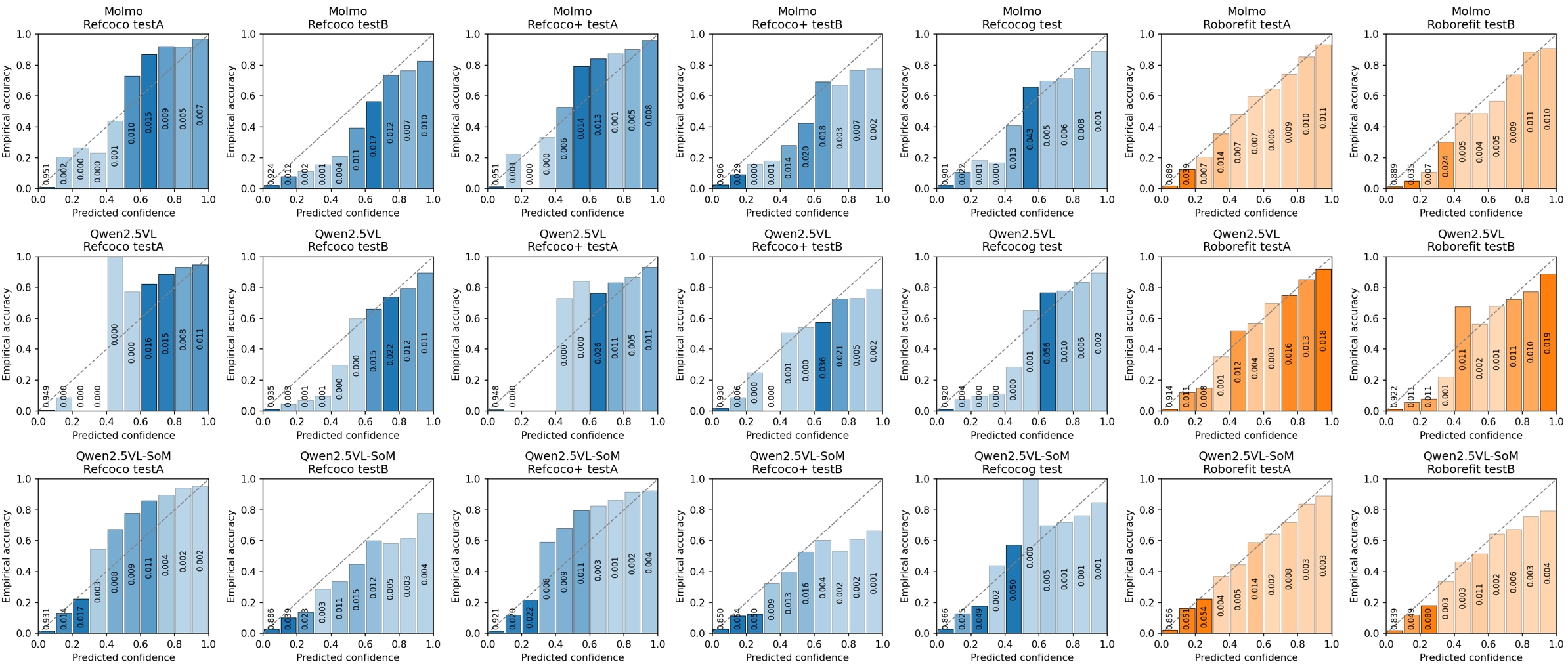}
        \label{fig:after_cal}
    }
    \caption{
    Reliability curves analysis. We validate our uncertainty estimation on multiple datasets and various types of visual grounding models, comparing (a) reliability diagrams of uncalibrated models, (b) calibrators from train (for RoboRefIt) or validation dataset (for RefCOCO series), and (c) reliability diagrams after post-hoc calibration (isotonic regression).
    }
    \label{fig:detailed_reliability_curve}
\end{figure}

\subsubsection{Detailed Results on Benchmarks}

We provide comprehensive evaluation results across three representative visual grounding model architectures: Molmo (pointing model), Qwen2.5VL (detection model), and Qwen2.5VL-SoM (scoring model). These models cover the three main paradigms discussed in \secref{sec:vg_models}.

\figref{fig:disc_detail} presents detailed discrimination metrics (AUROC, AUPRC, and Recall@K) across all model-dataset combinations, comparing three settings: \textit{Ours w/ Isotonic} with post-hoc calibration, \textit{Ours w/o Calibration} with uncertainty quantification only, and \textit{No Uncertainty} with raw model predictions.
Several key conclusions:
(1) Our uncertainty estimation consistently improves discrimination performance across all models, datasets, and metrics. The improvement is particularly substantial for AUPRC on harder test sets (e.g., RefCOCO testB, RefCOCO+ testB), confirming that calibrated uncertainty provides better ranking of correct predictions.
(2) Comparing across models, Qwen2.5VL achieves the highest baseline performance, while Qwen2.5VL-SoM benefits most from our uncertainty estimation. This suggests that our method is particularly effective for models with lower intrinsic confidence quality.
(3) The difference between Ours w/ Isotonic and Ours w/o Calibration is relatively small for discrimination metrics. This reveals that with quantized calibrated probabilities, the performance degradation is minimal.

\figref{fig:calib_detail} shows calibration metrics (ECE, aECE, 0.1$\times$NLL, Brier score) across all configurations, where lower values indicate better calibration.
With isotonic regression calibration, calibration errors are consistently reduced to a low level across all configurations, confirming the effectiveness of post-hoc calibration.
We can also see that our uncertainty quantification without post-hoc calibration already provides moderate improvement in calibration metrics, but the improvement is not as consistent as with post-hoc calibration.
Benefitting from larger held-out dataset, models trained on RoboRefIt show better calibration than RefCOCO series. This is consistent with the results in \figref{fig:data_scale_ablation_detail}.

\figref{fig:detailed_reliability_curve} provides a detailed analysis of reliability curves before and after calibration.
Before calibration (\figref{fig:no_cal}), all three model types exhibit significant miscalibration. Notably, Molmo and Qwen2.5VL predictions are mostly overconfident, i.e., most predictions concentrate at high confidence values but empirical accuracy is much lower, while Qwen2.5VL-SoM is underconfident in general.
After applying isotonic regression calibration (\figref{fig:after_cal}), the reliability curves align closely with the diagonal across all models and datasets, confirming that the post-hoc calibration method is effective to further improve the quality of uncertainty, resulting in more accurate and reliable probabilities.

\subsection{Additional Details of Interactive Visual Grounding}
\label{ap:ivg_baselines}

\subsubsection{Baselines}
\label{ap:ivg_baseline_details}

We compare our method against both uncertainty-free and uncertainty-aware policies to validate the effectiveness of our estimation. The baselines are designed to represent different assumptions about when and how to seek human confirmation during interactive visual grounding.

\noindent\textbf{Uncertainty-Free Baselines}:
These baselines do not use any explicit uncertainty estimation and instead rely on simple heuristics for query allocation.
\begin{itemize}
    \item \textbf{Confirm Randomly}: Queries objects uniformly at random without any rationale. This baseline serves as a lower bound for uninformed information gathering and establishes the baseline performance when query allocation is completely uninformed. It helps quantify the value of uncertainty-guided query strategies.
    \item \textbf{Confirm Every}: Verifies every detected object by asking a confirmation question for each. This represents the most conservative but highly inefficient strategy that ensures no target is missed at the cost of maximizing interaction burden. It provides an upper bound on achievable accuracy.
    \item \textbf{Confirm Positive}: A rule-based strategy that only verifies regions predicted as positive (i.e., predicted targets) by the visual grounding model. This simulates the typical verification heuristic used in prior interactive grounding systems~\citep{shridhar2018interactive}. It is motivated by the intuition that negative predictions are less likely to require verification.
\end{itemize}

\noindent\textbf{Uncertainty-Aware Baselines}:
These baselines incorporate some form of uncertainty estimation to guide confirmation decisions.
\begin{itemize}
    \item \textbf{VLM Self-Verbalization}: This baseline leverages a pretrained vision-language model (Qwen2.5VL-72B in our implementation) prompted to verbally articulate per-object confidence scores~\citep{linteaching}. Specifically, we prompt the VLM with each candidate object region and ask it to rate its confidence (a float ranging from 0 to 1) that the object matches the referring expression. The verbalized scores are then used to rank candidates. Objects with lower confidence are prioritized for confirmation queries. This baseline tests whether the implicit uncertainty awareness in large VLMs can substitute for explicit uncertainty estimation.
    \item \textbf{KnowNo}: This baseline applies conformal prediction~\citep{angelopoulos2021gentle} on top of VLM Self-Verbalization scores. Importantly, unlike the original KnowNo formulation~\citep{ren2023robots} which uses model logits, we use VLM Self-Verbalization scores because many state-of-the-art VLMs do not expose token-level logits for grounding tasks of many objects. The conformal calibration threshold is computed on a held-out set to achieve a target coverage rate. At test time, objects with scores below the threshold are included in the prediction set, and confirmation is triggered if the set contains more than one object. This baseline tests whether conformal prediction can rescue poorly calibrated VLM scores. We vary the target coverage rate to evaluate the trade-off between performance and efficiency.
\end{itemize}

\noindent\textbf{Our Conformal Policies}:
We integrate our estimated uncertainty with conformal prediction in two ways, leveraging the calibrated probabilities from our quantification and calibration pipeline. We include both because they target complementary operating regimes and isolate whether the gain comes from the conformal calibration alone or from a particular set-construction choice.
\begin{itemize}
    \item \textbf{Object-wise Conformal Policy}: Treats each object as an independent binary classification instance with the label set $\{$True, False$\}$. Using the calibrated $\tilde{p}_r$ as the positive-class score and $1-\tilde{p}_r$ as the negative-class score, the per-object conformal set is $\{$True$\}$ if $\tilde{p}_r \ge \tau(\alpha)$, $\{$False$\}$ if $1-\tilde{p}_r \ge \tau(\alpha)$, and ambiguous ($\{$True, False$\}$) when both classes are admissible. Confirmation is triggered only on ambiguity. We adopt this design because it gives a per-object coverage guarantee and avoids any top-$K$ truncation, so recall remains high when many candidates are plausible.
    \item \textbf{Image-wise Conformal Policy}: Treats the entire image as one instance and ranks all candidate objects by $\tilde{p}_r$, taking the top-$K$ ($K=5$) as the conformal prediction set and confirming sequentially through the ranked candidates. We adopt this design because, when the question budget is small, concentrating queries on the top few candidates yields the fastest accuracy ramp-up, at the cost of irrevocably dropping the target if it lies outside the top-$K$ set.
\end{itemize}

\subsubsection{Detailed Results}
\label{ap:ivg_result_details}

\begin{figure}[t]
    \centering
    \includegraphics[width=\textwidth]{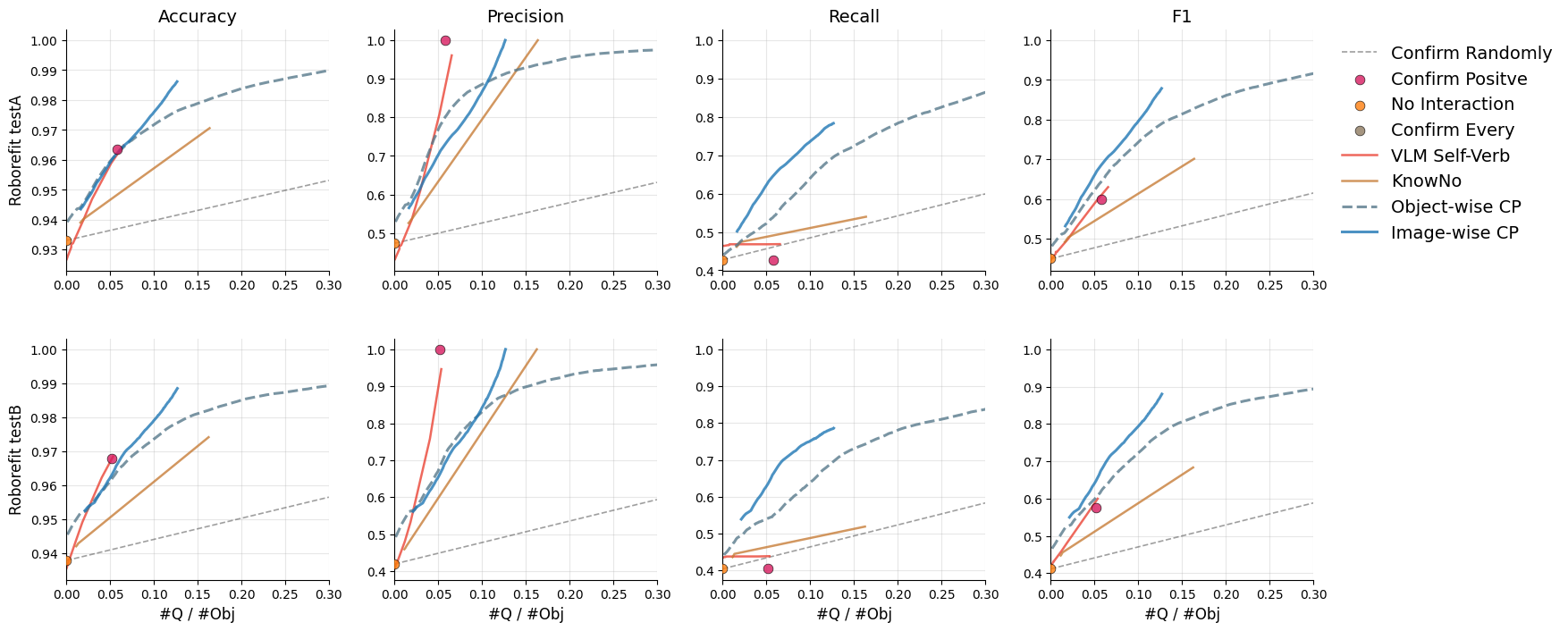}
    \caption{Interactive visual grounding performance on RoboRefIt test sets. We varied the decision boundary to control the question quota per object for each method, if applicable.}
    \label{fig:ivg}
\end{figure}

Results in \figref{fig:ivg} demonstrate that our conformal prediction-based policies achieve a superior trade-off between task performance and interaction cost.
\textit{Image-wise Conformal Prediction} rapidly reaches high accuracy with very few questions, making it ideal for extremely limited interaction budgets. However, its performance plateaus due to the fixed candidate set truncation.
In contrast, \textit{Object-wise Conformal Prediction} avoids truncation by evaluating each object independently, achieving higher recall and accuracy at the cost of slightly more questions.
Crucially, unlike heuristic thresholds, our conformal policies provide statistically grounded decision boundaries, allowing the system to target a specific risk level with formal guarantees.

A notable observation from \figref{fig:ivg} is that \textit{VLM Self-Verbalization} provides minimal improvement over the random query baseline, particularly in terms of recall. This can be attributed to several factors:
(1) VLMs tend to produce poorly calibrated confidence scores when prompted for self-assessment~\citep{linteaching, kadavath2022language}. The verbalized confidence often reflects the model's linguistic certainty rather than actual grounding accuracy.
(2) The confidence scores keep consistent and concentrate around the deterministic predictions. Hence, it produces similar confidence scores for multiple objects in the same image when given identical text queries, failing to distinguish between correct and incorrect predictions.
As a result, query ordering based on VLM Self-Verbalization scores offers little advantage over random selection.

Interestingly, \textit{KnowNo} quickly achieves high precision but its recall is improved incrementally compared to the random query baseline. The key reason lies in the candidate set construction:
In our implementation, KnowNo first filters candidates to those with non-zero VLM confidence scores, then selects the top-$K$ (where $K=5$) as the prediction set for conformal calibration.
When the number of non-zero confidence candidates is less than $K$, additional candidates are selected randomly to fill the set.
Since VLM Self-Verbalization often assigns low or zero confidence to many objects (due to poor calibration), a significant fraction of candidates in the KnowNo prediction set are effectively random.
This randomization inadvertently improves robustness by including objects that the VLM incorrectly dismissed, leading to performance that matches or slightly exceeds the random baseline.
However, this also means that KnowNo is limited by the VLM self-assessment capability, confirming that well-calibrated uncertainty estimates, as provided by our method, are essential for effective uncertainty-guided decision making.

\subsection{Additional Details of Robot Experiments}

This appendix details our real-robot evaluation: hardware and system stack, the object inventory, the two scene configurations, the baselines compared, and a per-mode failure analysis.

\begin{figure}[t]
    \centering
    \subfloat[Fetch robot setup.]{
        \includegraphics[width=0.48\linewidth]{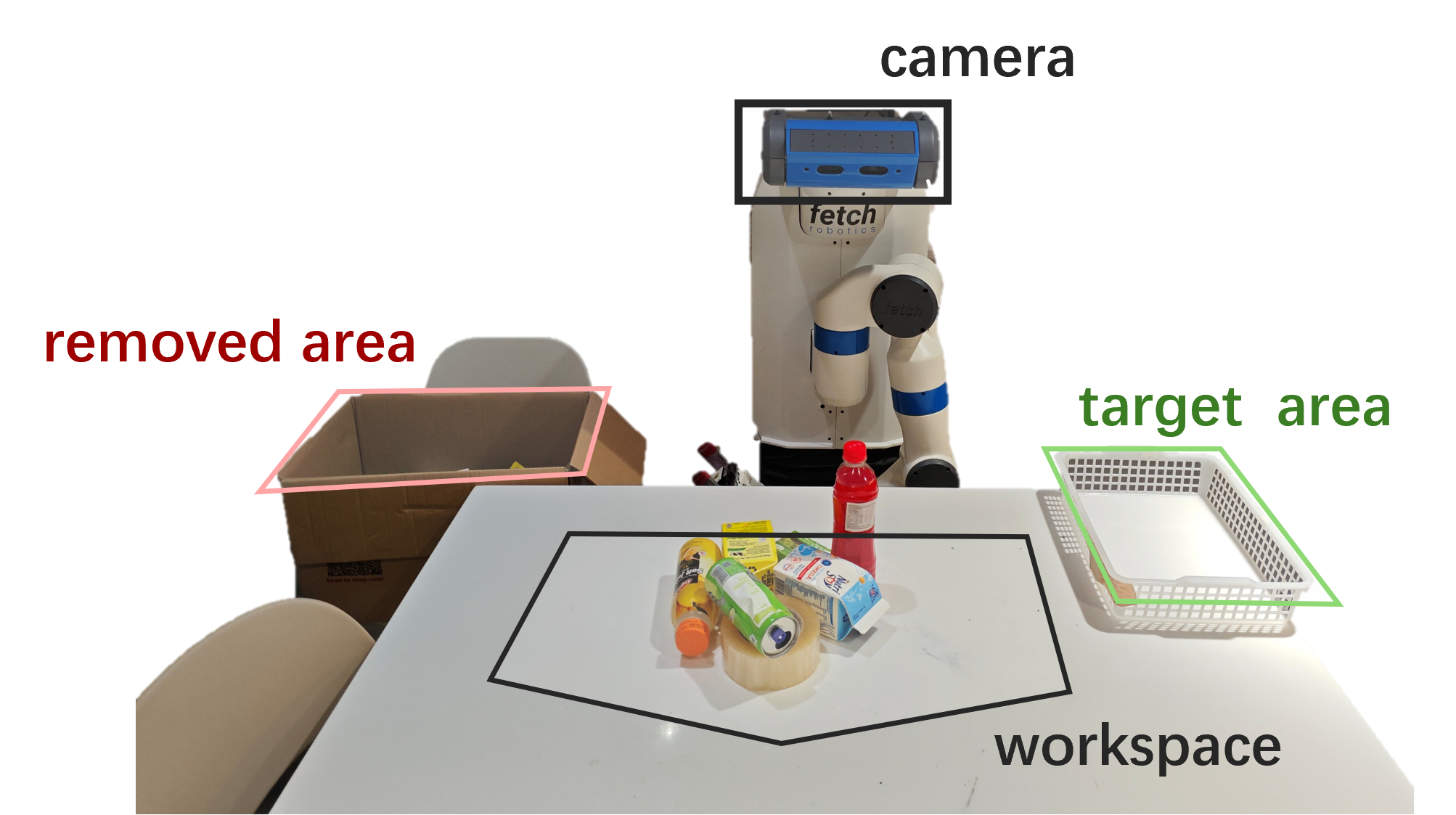}
        \label{fig:fetch}
    }
    \hfill
    \subfloat[Objects used in our experiments.]{
        \includegraphics[width=0.42\linewidth]{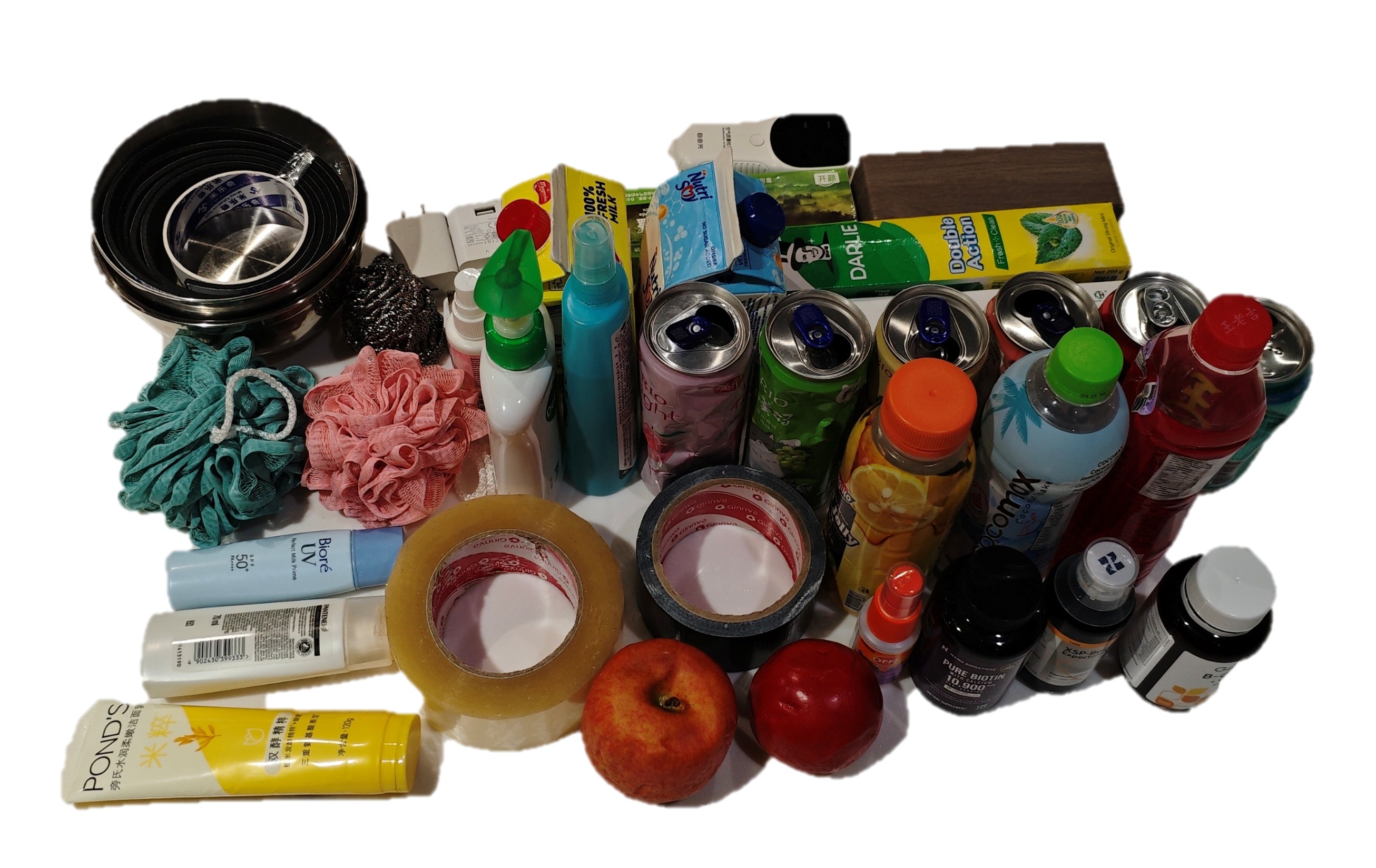}
        \label{fig:objects}
    }
    \caption{Real-robot experimental setup. (a) Fetch robot setup, which includes a 7-DoF arm for object manipulation, and a head camera for RGB-D perception. (b) All objects used in our real-robot experiments, which encompass different kinds of perceptual uncertainties. }
    \label{fig:robot_setup}
\end{figure}

\begin{figure}[t]
    \centering
    \includegraphics[width=\textwidth]{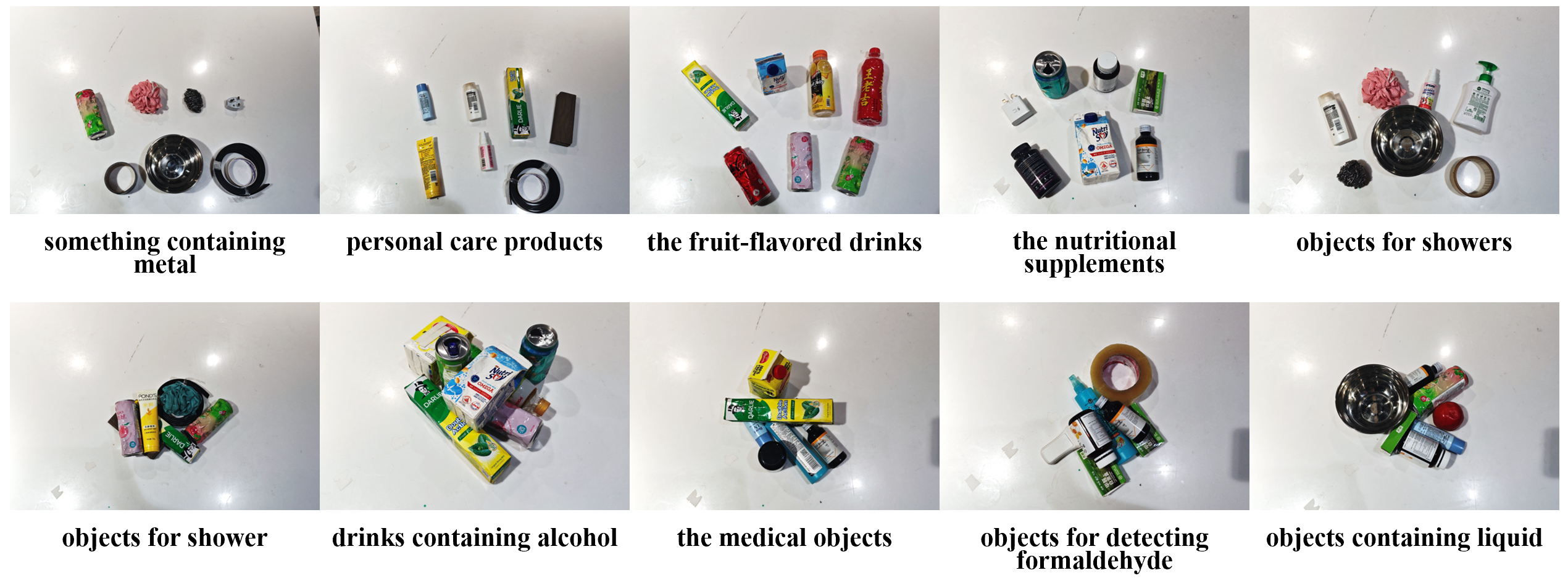}
    \caption{Robot experimental scenes. Top: 5 object-scattered scenes; Bottom: 5 object-stacked scenes. Instructions are illustrated under each image.}
    \label{fig:robot_scenes}
\end{figure}

\subsubsection{Hardware Setup} We illustrate the real-robot system setting in \figref{fig:robot_setup} used in our experiments.
We deploy a Fetch mobile manipulator equipped with a 7-DoF robotic arm for tabletop object manipulation and a head-mounted RGB-D camera that provides a nearly top-down view of the workspace for perception.
The workspace contains three functional regions: a tabletop manipulation workspace where objects are initially placed, a target area where selected objects that satisfy user instructions should be placed, and a removed area for objects that are excluded from the task.
The camera continuously observes the entire workspace, enabling visual grounding and state estimation during interaction.
A diverse set of everyday household objects with varying materials, functions, and visual appearances is used, intentionally introducing diverse kinds of perceptual uncertainties.
This setup allows us to systematically evaluate uncertainty-aware perception, decision making, and manipulation in a realistic cluttered tabletop environment.

\subsubsection{System Setup} Our system comprises different modules: perception modules, large-scale vision-language models, the uncertainty estimation module, and the hierarchical planning module, as introduced in \secref{sec:uncertainty} and \secref{sec:policy}.
For Qwen2.5VL-72B-AWQ \citep{bai2025qwen2} and Molmo-7B-D \citep{deitke2025molmo}, we run them with vLLM \citep{kwon2023efficient} for acceleration on two NVIDIA A100-80G GPUs.
For other perception models, we run them all together on another single A100-80G GPU, including the OWL-ViT \citep{minderer2023scaling} for object detection, Contact Graspnet \citep{sundermeyer2021contact} for grasp detection, VMRN \citep{zhang2018visual} for support relation detection, CLIP \citep{radford2021learning}, DINOv2 \citep{oquab2024dinov2}, and Sentence Transformer \citep{reimers2019sentence} for multi-modal embedding extraction and image-text pair retrieval.
Even though our system involves multiple vision-language models, a single visual-inference step over all modules after each physical grasp takes around 5-10 seconds.
The main cost is from calling the large-scale vision-language models, which depends on the input and output text lengths.
Other models are running with approximately real-time responses.
For efficient embedding retrieval, we make use of FAISS \citep{johnson2019billion}, which is a high-performance open-sourced toolkit to construct embedding databases and do fast retrieval.
For real-time motion planning, we use VAMP \citep{vamp_2024} with a fixed-base RRTC planner to generate collision-free robot arm paths.

Before each physical trial, the robot localizes itself using AMCL and navigates to a predefined manipulation base pose with respect to the tabletop workspace. After reaching this pose, the base is kept fixed during grasp execution. Given the target grasp pose, we first plan a collision-free arm trajectory to a pre-grasp configuration using a fixed-base RRTC planner in VAMP. The final approach to the grasp pose and the post-grasp lifting motion are generated as Cartesian-interpolated end-effector motions and converted into arm joint trajectories. All manipulation trajectories are executed by the default Fetch arm and torso trajectory-following controllers. For collision checking, we construct a point-cloud collision model from predefined workspace geometry and the current RGB-D observation, and use this model during arm motion planning.

\subsubsection{Objects for Robot Experiments}
We use a diverse set of everyday household objects with varying materials, shapes, functions, and visual appearances.
This includes rigid objects (bottles, boxes), deformable objects (cloth, bags), and transparent or reflective objects (glass, metal).
The diversity intentionally introduces various sources of perceptual uncertainty, such as visual ambiguity, occlusion, and lack of distinctive texture, to rigorously test the robustness of our uncertainty estimation and decision-making framework.

\subsubsection{Robot Experimental Scenes}
\label{ap:robot_scenes}

We evaluate active visual grounding in real-world settings, where the robot follows instructions to identify targets, optionally removing obstacles and asking questions to resolve uncertainty.
We report \textit{Success Rate} and the number of questions per scene (\textit{\#Q/\#Scene}) to measure effectiveness and efficiency.
We test two configurations to vary clutter and interaction difficulty. Each scene includes 7 objects, yielding 70 trials in total across the 10 scenes. Instructions are intentionally designed to include uncertainty.
\begin{itemize}
    \item \textbf{Scattered Scenes}: Objects are placed in a relatively sparse arrangement with minimal overlap. This tests the system's ability to handle basic visual grounding and geometric reasoning in a reachable workspace.
    \item \textbf{Stacked Scenes}: Objects are densely packed, physically touching, or stacked upon one another. This configuration introduces significant occlusion and physical obstruction, mimicking a ``bin picking'' scenario \citep{mahler2017learning}. It requires the robot to actively reason about blocking relationships, remove obstacles, and resolve high visual uncertainty through interaction.
\end{itemize}
We also illustrate all 10 scenes in \figref{fig:robot_scenes}.
We implement two different confirmation strategies:
\begin{itemize}
    \item \textit{Human Confirmation}: Confirmation is provided by a real human evaluator, who is assumed to be reliable. This isolates the system's planning and uncertainty-aware perception performance without introducing the errors from VLM confirmation. This setting is used in the main quantitative real-robot experiments.
    \item \textit{VLM Confirmation}: We implement an autonomous VLM-based confirmation mechanism. If an object is uncertain, the robot (1) picks up the candidate object, (2) presents it to the camera for a closer view, and (3) queries a VLM for confirmation. If the VLM remains uncertain or detects a mismatch, the robot proceeds to ask a human. This setting automates the confirmation process and allows more autonomy. This setting is used only for demonstration.
\end{itemize}

\subsubsection{Baselines of Robot Experiments}
\label{ap:robot_baselines}

We compare our planning-based method (\textbf{Ours-Qwen-P}, where ``P'' denotes the model-based planner of \apref{ap:planning}) against several strong baselines, each chosen to ablate a distinct part of the pipeline:

\begin{itemize}
    \item \textbf{GPT-5.1}: An end-to-end LLM policy that maps instructions and visual descriptions directly to actions, with no explicit uncertainty modeling and no specialized perception. We include it as the strongest off-the-shelf baseline, isolating what a state-of-the-art LLM does when given the task end-to-end.
    \item \textbf{Grounded GPT-5.1}: GPT-5.1 makes high-level decisions including when to ask while \emph{our} perception pipeline handles open-set detection, grounding, and grasping. We include it to isolate the value of our explicit uncertainty estimator versus an LLM-based controller, keeping perception pipeline fixed.
    \item \textbf{Ours-Qwen-H} and \textbf{Ours-GPT-H}: Heuristic variants of our system that replace the model-based planner with a simple threshold on the estimated uncertainty to decide whether to ask, otherwise greedily grasping or removing the most likely target. The two variants differ only in the question generator: \textbf{Ours-Qwen-H} uses Qwen2.5VL-72B to phrase the clarification question, while \textbf{Ours-GPT-H} uses GPT-5.1. By comparing with them, we can test whether the gains attributed to our uncertainty estimator hold when the planner is reduced to a heuristic policy, and whether the question-generator choice matters in practice.
\end{itemize}

\subsubsection{Detailed Failure Analysis}
\label{ap:fail_analysis}

\begin{table}[t]
\centering
\caption{Failure mode analysis across different methods.}
\label{tab:failures}
\begin{tabular}{lcccc}
\toprule
{\textbf{Method}} &
{\textbf{GRD}} &
{\textbf{DET}} &
{\textbf{GRP}} &
{\textbf{REL}} \\
\midrule
Ours-Qwen-P  & 8 & 1 & 1 & 0 \\
Ours-Qwen-H  & 9 & 2 & 0 & 0 \\
Ours-GPT-H  & 2 & 0 & 0 & 0 \\
Grounded GPT-5.1 & 28 & 1 & 2 & 0 \\
GPT-5.1        & 26 & 1 & 0 & 2 \\
\bottomrule
\end{tabular}
\end{table}

We observed different kinds of failure modes across the 70 real-robot trials per method.
Concretely, failures can be categorized into 4 classes:
\begin{itemize}
    \item \textbf{Grounding failures (GRD)} where failures are caused by associating the language instruction with the wrong objects.
    \item \textbf{Detection failures (DET)} where failures are from the false positives or false negatives of the object detector.
    \item \textbf{Grasping failures (GRP)} where occasionally, the robots may grasp the wrong object, which is not the grounded one, or grasp multiple objects at the same time due to dense clutter.
    \item \textbf{Relation failures (REL)} where the robot tries to grasp the objects that are blocked, and hence grasping fails. This usually happens for the GPT-5.1 baseline.
\end{itemize}
The failure counts are illustrated in \tabref{tab:failures}.
It shows that grounding failures (GRD) dominate across all methods and constitute the primary source of errors.
Detection (DET) and grasping (GRP) failures occur only occasionally and exhibit high variance, suggesting they are largely incidental rather than systematic.
In contrast, GRD errors account for the majority of failures, especially for GPT-5.1–based policies, originating from overconfident grounding decisions, even under highly uncertain situations.
Notably, GPT-5.1 occasionally misperceives blocking relations among objects, leading to incorrect target selection and grasping failure.
This further indicates that the main limitation lies in uncertainty handling during grounding and relational reasoning.

\subsection{Prompts}
\refstepcounter{prompt}\label{prompt:dialog_to_phrase}
\begin{tcolorbox}[
    colback=gray!5,
    colframe=black,
    boxrule=0.5pt,
    arc=2pt,
    left=4pt,
    right=4pt,
    top=4pt,
    bottom=4pt,
    breakable,
    title=\textbf{(\theprompt) Dialog-to-Phrase Summarization},
    fonttitle=\small\bfseries,
]
\footnotesize

\textbf{Task.} Summarize a dialog into a brief, precise phrase while preserving its original meaning.

If the input is a singular noun phrase, do be sure to make it PLURAL!
For example, use ``the cups on the table'' instead of ``the cup on the table''.

If the input does not specify the noun, make it a noun phrase.
For example, use ``something on top of the book'' instead of ``on top of the book''.

Please also fix the possible gramatical errors, if any.
For example, use ``point to the tubes with a faucet'' instead of ``point to tub has faucet''.

Do not miss important and meaningful information!

\vspace{0.3em}
\textbf{Example:}

\textbf{1. Dialog:}
\begin{verbatim}
on the left
\end{verbatim}
\textbf{Response:} something on the left

\vspace{0.3em}
\textbf{2. Dialog:}
\begin{verbatim}
Find the object I need.
Is it a kitchen item?
Yes.
Is it on the table?
Yes.
Is it a plate?
Yes.
\end{verbatim}
\textbf{Response:} The kitchen plates on the table.

\vspace{0.3em}
\textbf{3. Dialog:}
\begin{verbatim}
Give me the cup on the table.
Which one do you want?
The one on the left.
What color is it? Is it red or not?
Yes, it is red.
\end{verbatim}
\textbf{Response:} The red cups on the left side of the table.

\vspace{0.3em}
\textbf{4. Dialog:}
\begin{verbatim}
Something to cook.
Do you mean a pot?
No.
\end{verbatim}
\textbf{Response:} something to cook.

\vspace{0.3em}
\textbf{5. Dialog:}
\begin{verbatim}
{}
\end{verbatim}
\textbf{Response:}

\end{tcolorbox}

\refstepcounter{prompt}\label{prompt:informative_question}
\begin{tcolorbox}[
    colback=gray!5,
    colframe=black,
    boxrule=0.5pt,
    arc=2pt,
    left=4pt,
    right=4pt,
    top=4pt,
    bottom=4pt,
    breakable,
    title=\textbf{(\theprompt) Open-ended Question Generation},
    fonttitle=\small\bfseries,
]
\footnotesize

\vspace{0.3em}
\textbf{Task.} You are skilled in sorting objects with limited information. Your goal is to identify the target object(s) efficiently. Given an image and dialogue history, generate a precise question to reduce any uncertainty or ambiguity and identify the target object(s).
You need to generate the MOST informative question with possible answer candidates to gather as much information as possible.
Do not directly include object IDs in the generated question.
If applicable, the objects to be confirmed are labeled with unique object IDs with masks.

\vspace{0.3em}
\textbf{Instructions.}

\textbf{1. Question Generation:}

Question must be concise, targeted, and systematically reduce uncertainty or ambiguity (e.g., color, shape, material, size, location, function).
Avoid redundancy and overly broad questions.
It should be the most informative one, directly contributing to object identification.

\vspace{0.3em}
\textbf{2. Answer Candidates}

Predict realistic user responses for the question based on all language and visual contexts.
Do not provide non-specific or vague answers like ``others'', ``unknown'' for open-ended questions.

\vspace{0.3em}
\textbf{Your response must be a valid JSON dict with the following keys:}
\begin{itemize}\setlength{\itemsep}{0pt}
    \item \texttt{"question"}: A string representing the question.
    \item \texttt{"answers"}: A list of candidate answer strings.
\end{itemize}

\vspace{0.3em}
\textbf{Example 1:}
\begin{verbatim}
{
  "question": "What color is it?",
  "answers": ["Red", "Yellow", "Black"]
}
\end{verbatim}

\textbf{Example 2:}
\begin{verbatim}
{
  "question": "Where are they located?",
  "answers": [
    "On the table", 
    "Scattered on the floor", 
    "Next to the lamp"
  ]
}
\end{verbatim}

\textbf{Example 3:}
\begin{verbatim}
{
  "question": "which one do you mean?",
  "answers": [
    "The one on the left", 
    "The one with a cap", 
    "All of them, please"
  ]
}
\end{verbatim}

\textbf{Example 4:}
\begin{verbatim}
{
  "question": "Is it made of metal?",
  "answers": ["Yes", "No"]
}
\end{verbatim}

\textbf{Example 5:}
\begin{verbatim}
{
  "question": "What size are they?",
  "answers": [
    "Small", 
    "A mix of small and large",
    "Large"
  ]
}
\end{verbatim}

\textbf{Example 6:}
\begin{verbatim}
{
  "question": "Do you mean the bags?",
  "answers": [
    "Yes", 
    "No, I meant the boxes"
  ]
}
\end{verbatim}

\end{tcolorbox}

\refstepcounter{prompt}\label{prompt:action_decision}
\begin{tcolorbox}[
    colback=gray!5,
    colframe=black,
    boxrule=0.5pt,
    arc=2pt,
    left=4pt,
    right=4pt,
    top=4pt,
    bottom=4pt,
    breakable,
    title=\textbf{(\theprompt) VLM as the Policy},
    fonttitle=\small\bfseries,
]
\footnotesize

\textbf{Task.} You are an expert in object search and decision-making.
Given an image as your visual observation, your task is to identify and interact with a specified set of target objects.
Each object is highlighted with a unique color and numeric OBJECT\_ID.

\vspace{0.3em}
\textbf{Available actions:}
\begin{itemize}\setlength{\itemsep}{0pt}
    \item PickUp – Select and pick up a correctly identified target object.
    \item Remove – Eliminate an object from consideration if it is determined to be incorrect or undesired.
    \item Ask – Pose a specific, open-ended question to reduce uncertainty before taking action.
    \item End - If you think that the task is fully finished, generate an ``End'' action to terminate.
\end{itemize}

\vspace{0.3em}
\textbf{Output format:}
Respond with a JSON list containing 4 elements:
\begin{enumerate}\setlength{\itemsep}{0pt}
    \item your short reason for choosing the specific action to execute, and please explain yourself first.
    \item the action type from \{\texttt{Ask}, \texttt{Remove}, \texttt{PickUp}, \texttt{End}\}.
    \item the object id, if applicable, else null.
    \item the question, if the action is ``Ask'', else null.
\end{enumerate}

\vspace{0.3em}
\textbf{Requirements:}
\begin{itemize}\setlength{\itemsep}{0pt}
    \item Please note that you are now in process of the task, there might be no target objects satisfying the description.
    \item If you are sure that no objects are targets, remove all the rest please.
    \item If you are certain about the identity and relevance of your targets, take action immediately.
    \item If you are uncertain in perception or task requirements, ask a question to clarify before proceeding.
    \item The question should be the most informative one, directly contributing to object identification.
    \item The question should not involve complex reasoning or expertise knowledge.
    \item Avoid generating repetitive questions.
    \item Do not directly include object IDs in the generated question.
    \item Avoid asking questions like ``does the highlighted objects ...'', since user cannot see the image and object IDs.
    \item You must decide for each object until there is not objects that are near you and graspable.
\end{itemize}

\vspace{0.3em}
\textbf{A Running Example:}

\textbf{An example where the instruction is highly UNCERTAIN:}
\begin{verbatim}
User: Hand me the book.
Assistant: [
  "<YOUR_SHORT_REASON>", 
  "Ask", null,
  "Which book do you want?"
]
\end{verbatim}

\textbf{An example where object OBJECT\_ID is UNCERTAIN to be the target:}
\begin{verbatim}
User: Hand me the book.
Assistant: Which book do you want?
User: The blue-colored.
Assistant: [
  "<YOUR_SHORT_REASON>", 
  "Ask", OBJECT_ID,
  "Do you mean the notebook on the table?"
]
\end{verbatim}

\textbf{An example where object OBJECT\_ID is NOT THE TARGET, particularly when it is OBSTRUCTING THE TARGET:}
\begin{verbatim}
User: Hand me the book.
Assistant: Which book do you want?
User: The blue-colored one.
Assistant: Do you mean the notebook on the table?
User: Yes, the one on the table.
Assistant: [
  "<YOUR_SHORT_REASON>", 
  "Remove", OBJECT_ID, 
  null
]
\end{verbatim}

\textbf{An example where the TARGET IS CLEARLY VISIBLE AND CAN BE GRASPED DIRECTLY:}
\begin{verbatim}
User: Hand me the book.
Assistant: Which book do you want?
User: The blue-colored one.
Assistant: Do you mean the notebook on the table?
User: All of them, please.
Assistant: [
  "<YOUR_SHORT_REASON>", 
  "PickUp", OBJECT_ID, null
]
\end{verbatim}

Your decisions should be based on the available observations and dialogues to ensure correct object selection.
\end{tcolorbox}

\refstepcounter{prompt}\label{prompt:action_decision_grounded}
\begin{tcolorbox}[
    colback=gray!5,
    colframe=black,
    boxrule=0.5pt,
    arc=2pt,
    left=4pt,
    right=4pt,
    top=4pt,
    bottom=4pt,
    breakable,
    title=\textbf{(\theprompt) Decision-making with Grounding Models},
    fonttitle=\small\bfseries,
]
\footnotesize

\textbf{Task.} You are an expert in object search and decision-making. Given an image as your visual observation, your task is to identify and interact with a specified set of target objects.
Each object is highlighted with a unique color and numeric OBJECT\_ID.

\vspace{0.3em}
\textbf{Available actions:}

PickUp – Select and pick up a correctly identified target object.

Ask – Pose a specific, open-ended question to reduce uncertainty before taking action.

End - If you think that the task is fully finished, generate an ``End'' action to terminate.

\vspace{0.3em}
\textbf{Output format:}

Respond with a JSON list containing 4 elements:
\begin{enumerate}\setlength{\itemsep}{0pt}
    \item your short reason for choosing the specific action to execute, and please explain yourself first.
    \item the action type from \{\texttt{Ask}, \texttt{PickUp}, \texttt{End}\}.
    \item the object id, if applicable, else null.
    \item the question, if the action is ``Ask'', else null.
\end{enumerate}

\vspace{0.3em}
\textbf{Requirements:}
\begin{itemize}\setlength{\itemsep}{0pt}
    \item Please note that you are now in process of the task, there might be no target objects satisfying the description.
    \item If you are sure that no objects are targets, please end the task.
    \item If you are certain about the identity and relevance of your targets, take action immediately.
    \item If you are uncertain in perception or task requirements, ask a question to clarify before proceeding.
    \item The question should be the most informative one, directly contributing to object identification.
    \item The question should not involve complex reasoning or expertise knowledge.
    \item Avoid generating repetitive questions.
    \item Do not directly include object IDs in the generated question.
    \item Avoid asking questions like ``does the highlighted objects ...'', since user cannot see the image and object IDs.
\end{itemize}

\vspace{0.3em}
\textbf{Examples:}

\textbf{An example where the instruction is highly UNCERTAIN:}
\begin{verbatim}
User: Hand me the book.
Assistant: [
  "<YOUR_SHORT_REASON>", 
  "Ask", null,
  "Which book do you want?"
]
\end{verbatim}

\textbf{An example where object OBJECT\_ID is UNCERTAIN to be the target:}
\begin{verbatim}
User: Hand me the book.
Assistant: Which book do you want?
User: The blue-colored.
Assistant: [
  "<YOUR_SHORT_REASON>", 
  "Ask", OBJECT_ID,
  "Do you mean the notebook on the table?"
]
\end{verbatim}

\textbf{An example where THE TARGET IS CLEARLY VISIBLE:}
\begin{verbatim}
User: Hand me the book.
Assistant: Which book do you want?
User: The blue-colored one.
Assistant: Do you mean the notebook on the table?
User: All of them, please.
Assistant: [
  "<YOUR_SHORT_REASON>", 
  "PickUp", OBJECT_ID, null
]
\end{verbatim}

\textbf{An example where NO TARGET exists:}
\begin{verbatim}
User: Hand me the book.
Assistant: Which book do you want?
User: The blue-colored one.
Assistant: Do you mean the notebook on the table?
User: All of them, please.
Assistant: [
  "<YOUR_SHORT_REASON>", 
  "PickUp", OBJECT_ID, null
]
\end{verbatim}

Your decisions should be based on the available observations and dialogues to ensure correct object selection.
\end{tcolorbox}

\refstepcounter{prompt}\label{prompt:som}
\begin{tcolorbox}[
    colback=gray!5,
    colframe=black,
    boxrule=0.5pt,
    arc=2pt,
    left=4pt,
    right=4pt,
    top=4pt,
    bottom=4pt,
    breakable,
    title=\textbf{(\theprompt) Set of Masks Prompt},
    fonttitle=\small\bfseries,
]
\footnotesize

\textbf{Task.} You are given an image and a set of candidate object regions.
Each region is represented by:
\begin{itemize}\setlength{\itemsep}{0pt}
    \item an ID (mask\_i),
    \item a segmentation mask already overlapped on the image.
\end{itemize}

\vspace{0.3em}
Your task: select the regions that match the referring expression.

\vspace{0.3em}
\textbf{Referring expression:} \texttt{"{}"}

\vspace{0.3em}
\textbf{Instructions:}
\begin{enumerate}\setlength{\itemsep}{0pt}
    \item Carefully read the referring expression.
    \item Compare it against the visual and spatial attributes of each candidate region.
    \item Choose the matched regions.
    \item Output in the following JSON format:
\end{enumerate}

\begin{verbatim}
{
  "region_ids": [<MASK_ID1>, ...], 
  "reason": "<short explanation of why these 
    regions match the referring expression>"
}
\end{verbatim}

\end{tcolorbox}

\refstepcounter{prompt}\label{prompt:lexical_rewrite}
\begin{tcolorbox}[
    colback=gray!5,
    colframe=black,
    boxrule=0.5pt,
    arc=2pt,
    left=4pt,
    right=4pt,
    top=4pt,
    bottom=4pt,
    breakable,
    title=\textbf{(\theprompt) Instruction Rephrasing},
    fonttitle=\small\bfseries,
]
\footnotesize

\textbf{Task.} You are given a short referring expression or instruction.

Rewrite the sentence using LESS COMMON words, while preserving the original meaning and referring to the exact same target object(s).

\vspace{0.3em}
\textbf{Constraints:}
\begin{itemize}\setlength{\itemsep}{0pt}
    \item Replace or rephrase \texttt{\{NWORD\}} words. Note that you can replace the word with a phrase if they share the same meaning.
    \item Be make sure that the resulting words are with the same meaning.
    \item Use rarer or less usual synonyms when possible, but keep the sentence fluent and natural.
    \item Do NOT add new attributes, relations, or quantities.
\end{itemize}

\vspace{0.3em}
\textbf{Original sentence:}

\texttt{"\{SENTENCE\}"}

\vspace{0.3em}
Return only the rewritten sentence.

\end{tcolorbox}

\refstepcounter{prompt}\label{prompt:bbox_grounding}
\begin{tcolorbox}[
    colback=gray!5,
    colframe=black,
    boxrule=0.5pt,
    arc=2pt,
    left=4pt,
    right=4pt,
    top=4pt,
    bottom=4pt,
    breakable,
    title=\textbf{(\theprompt) Qwen2.5VL Visual Grounding},
    fonttitle=\small\bfseries,
]
\footnotesize

\textbf{Task.} Please find out all matched objects and bounding boxes given the input referring expression:
\texttt{\{\}}

\vspace{0.3em}
The output format should be like a list of bounding boxes.
Do not include any other texts except for the numbers.

\vspace{0.3em}
\textbf{Output format:}
\begin{verbatim}
[
    [X1, Y1, X2, Y2], 
    [X1, Y1, X2, Y2], 
    ...
]
\end{verbatim}

\end{tcolorbox}

\refstepcounter{prompt}\label{prompt:knowno}
\begin{tcolorbox}[
    colback=gray!5,
    colframe=black,
    boxrule=0.5pt,
    arc=2pt,
    left=4pt,
    right=4pt,
    top=4pt,
    bottom=4pt,
    breakable,
    title=\textbf{(\theprompt) KnowNo Prompting-based Uncertainty},
    fonttitle=\small\bfseries,
]
\footnotesize

\textbf{Task.} You are given an image with labeled object IDs and a referring expression: \texttt{\{\}}.
Your task is to identify the possible matched object or objects in the scene and assign each a confidence score between 0 and 1.
Please make sure to include every object that is potentially relevant to the referring expression.
Do not include explanations or additional text outside the JSON list.

\vspace{0.3em}
Return the results as a simple JSON list of dictionaries in the following format:
\begin{verbatim}
[
  {
    "region_id": <MASK_ID1>,
    "score": <CONFIDENCE_SCORE1>,
    "reason": <short explanation of why>
  },
  ...
]
\end{verbatim}

\end{tcolorbox}

\end{document}